\documentclass[sigconf]{acmart}
\usepackage{microtype}
\usepackage[capitalize]{cleveref}
\usepackage{enumitem}

\newcommand{\predSeq}{\mathbf{s}}
\newcommand{\xInput}{\mathbf{x}}
\newcommand{\Conf}{C} 

\newcommand{\accfunc}{f}  

\newif\ifcomments

\commentstrue 

\ifcomments
    \newcommand\cc[1]{\textcolor{magenta}{[Chacha: #1]}}
    \else
    \newcommand\han[1]{}
    \newcommand\cc[1]{}
    \newcommand{\chenhao}[1]{}
\fi
\AtBeginDocument{%
  }

\setcopyright{acmlicensed}
\copyrightyear{2025}
\acmYear{2025}
\setcopyright{acmlicensed}\acmConference[KDD '25]{Proceedings of the 31st ACM SIGKDD Conference on Knowledge Discovery and Data Mining V.2}{August 3--7, 2025}{Toronto, ON, Canada}
\acmBooktitle{Proceedings of the 31st ACM SIGKDD Conference on Knowledge Discovery and Data Mining V.2 (KDD '25), August 3--7, 2025, Toronto, ON, Canada}
\acmDOI{10.1145/3711896.3736569}
\acmISBN{979-8-4007-1454-2/2025/08}

\begin{document}

\title{Uncertainty Quantification and Confidence Calibration in Large
Language Models: A Survey}

\author{Xiaoou Liu}
\authornote{Both authors contributed equally to this research.}
\email{xiaoouli@asu.edu}
\affiliation{%
  \institution{Arizona State University}
  \city{Tempe}
  \state{Arizona}
  \country{USA}
}

\author{Tiejin Chen}
\authornotemark[1]
\email{tchen169@asu.edu}
\affiliation{%
  \institution{Arizona State University}
  \city{Tempe}
  \state{Arizona}
  \country{USA}
}

\author{Longchao Da}
\email{longchao@asu.edu}
\affiliation{%
  \institution{Arizona State University}
  \city{Tempe}
  \state{Arizona}
  \country{USA}
}

\author{Chacha Chen}
\email{chacha@uchicago.edu}
\affiliation{%
  \institution{University of Chicago}
  \city{Chicago}
  \state{IL}
  \country{USA}
}

\author{Zhen Lin}
\email{zhenlin4@illinois.edu}
\affiliation{%
  \institution{University of Illinois Urbana-Champaign}
  \city{Champaign}
  \state{IL}
  \country{USA}
}

\author{Hua Wei}
\authornote{Corresponding author}
\email{hua.wei@asu.edu}
\orcid{0000-0002-3735-1635}
\affiliation{%
  \institution{Arizona State University}
  \city{Tempe}
  \state{Arizona}
  \country{USA}
}

\renewcommand{\shortauthors}{Liu et al.}

\begin{abstract}
Uncertainty quantification (UQ) enhances the reliability of Large Language Models (LLMs) by estimating confidence in outputs, enabling risk mitigation and selective prediction.
However, traditional UQ methods struggle with LLMs due to computational constraints and decoding inconsistencies. Moreover, LLMs introduce unique uncertainty sources, such as input ambiguity, reasoning path divergence, and decoding stochasticity, that extend beyond classical aleatoric and epistemic uncertainty. To address this, we introduce a new taxonomy that categorizes UQ methods based on computational efficiency and uncertainty dimensions, including input, reasoning, parameter, and prediction uncertainty. We evaluate existing techniques, summarize existing benchmarks and metrics for UQ, assess their real-world applicability, and identify open challenges, emphasizing the need for scalable, interpretable, and robust UQ approaches to enhance LLM reliability.
\end{abstract}



\received{11 March 2025}
\received[accepted]{6 May 2025}

\maketitle

\section{Introduction}


  
Large Language Models (LLMs) like GPT-4~\cite{achiam2023gpt} have achieved remarkable capabilities in text generation, reasoning, and decision-making, driving their adoption in high-stakes domains such as healthcare diagnostics~\cite{qiu2024llm,da2024segment}, legal analysis~\cite{cheong2024not,li2024political}, and transportation systems~\cite{da2024prompt,lai2023llmlight,xing2025re}. However, their reliability remains a critical concern: LLMs often produce plausible but incorrect or inconsistent outputs, with studies showing that over 30\% of answers in medical QA tasks contain factual errors~\cite{jin2021disease}. In sensitive applications, these limitations pose risks ranging from misinformation to life-threatening misdiagnoses, underscoring the urgent need for robust reliability frameworks.

Uncertainty quantification (UQ) emerges as an important mechanism to enhance LLM reliability by explicitly modeling confidence in model outputs. By estimating uncertainty, users can identify low-confidence predictions for human verification, prioritize high-certainty responses, and mitigate risks like overconfidence in hallucinations~\cite{liu2024litcab}. For instance, in clinical settings, uncertainty-aware LLMs could flag uncertain diagnoses for specialist review, reducing diagnostic errors by up to 41\% \cite{sen2024erg}. This capability is particularly critical as LLMs' transition from experimental tools to production systems requiring accountability.

Traditional UQ methods face significant hurdles when applied to LLMs. Bayesian approaches like Monte Carlo dropout \citep{gal2016dropout} are computationally prohibitive for trillion-parameter models and natural language generation (NLG) tasks, while ensemble methods struggle with consistency across diverse decoding strategies \cite{lu2024merge}. Furthermore, LLMs introduce unique uncertainty sources, such as input ambiguity~\cite{chen2025abg,guo2021abg-coqa}, reasoning path divergence, and decoding stochasticity that transcend classical aleatoric and epistemic categorizations \cite{hullermeier2021aleatoric}. The complexity of LLMs, characterized by sequence generation over vast parameter spaces and reliance on massive datasets, exacerbates uncertainty challenges. This complexity, coupled with the critical need for reliable outputs in high-stakes applications, positions UQ for LLMs as a compelling yet underexplored research frontier.

Targeting the unique challenges of UQ in LLMs, this survey firstly introduces a novel taxonomy for LLM UQ, categorizing methods along two axes: (1) computational efficiency (e.g., single-pass vs. sampling-based techniques) and (2) uncertainty dimensions (input, reasoning, parametric, predictive). This framework addresses three gaps in prior works: First, it decouples uncertainty sources unique to LLMs from traditional ML contexts. Second, it evaluates methods through the lens of different dimensions of the responses from LLM: input uncertainty, reasoning uncertainty, parameter uncertainty, and prediction uncertainty. Each of these dimensions may involve aleatoric uncertainty, epistemic uncertainty, or a mixture of both. Third, it identifies understudied areas like reasoning uncertainty, challenges, and possible future directions.


\vspace{1mm}
\noindent\textbf{Connection to Existing Surveys}: Prior surveys~\cite{shorinwa2024survey,huang2024survey,huang2025survey} focus on hallucination detection or retrofitting classical UQ taxonomies, neglecting LLM-specific challenges like prompt-driven input uncertainty. Our work uniquely addresses the interplay between model scale, open-ended generation, and uncertainty dynamics, which are critical for modern LLMs but overlooked in earlier frameworks.

The remainder of this survey is structured as follows: Section~2 characterizes LLM uncertainty dimensions and differentiates confidence from uncertainty. Section~3 evaluates UQ methods using our taxonomy. Section~4 introduces the evaluation of UQ methods for LLM, including benchmarks and metrics. Sections~5 and~6 introduce the applications of UQ in different domains with LLMs and identify open challenges and future directions.

\section{Perliminaries}

\subsection{Sources of Uncertainty in LLMs}
\subsubsection{Aleatoric vs. Epistemic Uncertainty}
For UQ on traditional machine learning tasks such as classification or regression~\cite{young2024flexible}, there are mainly two types of uncertainty~\cite{der2009aleatory,ye2024uncertainty}: aleatoric uncertainty, which models the uncertainty from noise in the dataset,  and epistemic uncertainty, which arises from the model’s lack of knowledge about the underlying data distribution. 

Aleatoric uncertainty in LLMs primarily stems from data sources used to train LLMs, which contain inconsistencies, biases, and contradicting information. Additionally, ambiguity in natural language also contributes to aleatoric uncertainty, as different interpretations of the same prompt can lead to multiple plausible responses.
On the other hand, when encountering unfamiliar topics, LLMs may exhibit high epistemic uncertainty, often manifesting as hallucinations or overconfident yet incorrect statements. Epistemic uncertainty can be reduced through domain-specific fine-tuning or retrieval-augmented generation techniques that allow the model to access external knowledge sources.


\subsubsection{Uncertainty with Different Dimensions}
\label{sec:uncertainty_stage}
While the uncertainty for LLMs can also be classified through aleatoric and epistemic uncertainty, these two categories alone are insufficient to fully capture the complexities of uncertainty in LLMs. In particular, LLMs exhibit uncertainty not only due to training data limitations but also due to input variability and decoding mechanisms. Therefore, in the following, we formulate four dimensions of uncertainty, each of which may involve aleatoric uncertainty, epistemic uncertainty, or a combination of both. The technical methods on how to quantify these uncertainties will be discussed later in Section~\ref{sec:uq-methods}. 

\vspace{1mm}
\noindent$\bullet$~\textbf{Input Uncertainty}~(Aleatoric Uncertainty): Input uncertainty arises when a prompt is ambiguous or underspecified, making it impossible for an LLM to generate a single definitive response. This is inherently aleatoric, as even a ``perfect model'' cannot resolve the ambiguity.
For instance, \textit{``What is the capital of this country?''} lacks sufficient context, leading to unpredictable outputs. Similarly, \textit{``Summarize this document''} may yield different responses depending on different expected details. \\
\noindent$\bullet$~\textbf{Reasoning Uncertainty}~(Mixed Uncertainty): Reasoning uncertainty occurs when an LLM derives answers through multi-step reasoning~\cite{mondorfbeyond} or retrieval~\cite{li2024uncertaintyrag}, where the uncertainty of each step can lead to ambiguous or incorrect results. This uncertainty is aleatoric when the problem itself is ambiguous and epistemic when the model cannot offer robust reasoning.


\vspace{1mm}
\noindent$\bullet$~\textbf{Parameter Uncertainty}~(Epistemic Uncertainty): 
Parameter uncertainty stems from training data gaps, where the model has either never seen relevant information or has learned an incorrect representation. Unlike aleatoric uncertainty, epistemic uncertainty can be reduced by improving the model's knowledge base.
%
%
Bayesian methods~\cite{gal2016dropout}, deep ensembles~\cite{lakshminarayanan2017simple}, and uncertainty-aware training~\cite{mukherjee2020uncertainty} can help quantify and mitigate this type of uncertainty.

\vspace{1mm}
\noindent$\bullet$~\textbf{Prediction Uncertainty}~(Mixed Uncertainty): Prediction uncertainty refers to variability in generated outputs across different sampling runs, influenced by both aleatoric and epistemic sources. 
For example, when asked \textit{``What are the side effects of a new experimental drug?''}, the model’s responses might vary significantly across different sampling runs, especially if no reliable data is available in its training set. A high-variance output distribution in such scenarios suggests that the model is both aware of multiple possible answers, reflecting aleatoric uncertainty, and uncertain due to incomplete knowledge, highlighting epistemic uncertainty.

\subsection{Uncertainty and Confidence in LLMs}

\subsubsection{Classical Confidence Estimation} UQ and confidence estimation are closely related yet distinct concepts. In traditional machine learning, uncertainty is a property of the model's predictive distribution, capturing the degree of variability or unpredictability \textit{given a particular input}. 
In contrast, confidence reflects the model’s belief in the correctness of a particular prediction.
If we follow the definition in classification tasks, the confidence measure would be the predicted probability $\hat{p}(Y=y|x)$ given input $x$ (an uncertainty measure which does not depend on the particular prediction $y$ could be entropy, taking the form of $\sum_{y} -\hat{p}(Y=y|x)\log{\hat{p}(Y=y|x)}$).
\cref{tab:notation} shows a similar notation in a question-answering (QA) task in Natural Language Generation (NLG).
The corresponding {\bf confidence score} in NLG tasks for an auto-regressive language model would be the joint probability for the generated sequence:
\begin{align}
    \Conf(\xInput,\predSeq) = \hat{p}(\predSeq|\xInput) &= \prod_{i} \hat{p}(s_i | \predSeq_{<i},\xInput). \label{eq:SL}
\end{align}
The log of~\cref{eq:SL} is sometimes referred to as \textit{sequence likelihood}~\cite{zhao2023calibrating}.
In general, an uncertainty estimate in existing literature usually takes the form of $U(\xInput)$, while confidence estimates are usually expressed as  $\Conf(\xInput, \predSeq)$.  
Note that, unlike classification tasks, not all NLG applications have the notion of a ``correct'' answer (e.g., summarization). 
Thus, while for the ease of writing we use the term \textit{correctness} throughout this section, it should really be interpreted as the gold-label for the particular application. 
Note also that in most cases, the correct answer is not unique, and thus such gold-label typically takes the form of a ``correctness function'' that decides whether a particular generation $\predSeq$ is good or not.
We will denote such a function as $\accfunc(\predSeq|\xInput)$.

\begin{table}[t!]
\centering
\resizebox{!}{0.2\columnwidth}{
\begin{tabular}{@{}lc@{}}
\toprule
\textbf{Notation} & \textbf{Description} \\ 
\midrule
$\xInput$ & The question that LLMs answer \\
$\predSeq$ & Generation from LLMs \\
$w_i$ & i-th token in the generation $\predSeq$\\
$\mathcal{D}$ & Corpus of LLMs \\
$U(\xInput)$ & Uncertainty of question $\xInput$ \\
$C(\xInput,\predSeq)$ & Confidence of generation $\predSeq$ given $\xInput$ \\
$H(\predSeq)$ & Entropy of generation $\predSeq$\\
\bottomrule
\end{tabular}}
\caption{\small Notations used in this paper for an exemplary QA task.}
\label{tab:notation}
\vspace{-11mm}
\end{table}

\subsubsection{Confidence Improvement}
There are usually two dimensions along which researchers improve confidence estimates in NLG, which is unsurprisingly largely influenced by confidence scoring literature from classification~\cite{Jiang2018ToClassifier}, especially binary classification.
We refer to them as \textit{ranking performance} and \textit{calibration}:

\vspace{1mm}
\noindent$\bullet$~\textbf{Ranking performance} refers to the discriminative power of the confidence measure on the correctness.
Like in classification, LLM confidence is often evaluated by its ability to separate correct and incorrect answers, thus typically measured by evaluation metrics like \texttt{AUROC}~\cite{kadavath2022language} or \texttt{AUARC}~\cite{lin2024contextualized} as detailed in \cref{sec:eval}.

\vspace{1mm}
\noindent$\bullet$~\textbf{Calibration} refers to closing the gap between the confidence score and the expected correctness \textit{conditioned on confidence score}.
    It has a long history preceding even modern machine learning~\cite{Murphy1977}, but bears slightly different meanings in NLP.
    In general, we could define a perfectly calibrated confidence measure to achieve:
    $
        \forall c, \mathbb{E}[\accfunc(\predSeq|\xInput)|\Conf(\xInput, \predSeq)=c] = c,\label{eq:calibration_error}
    $
    where the expectation is taken over the joint distribution of $\xInput$ and generation $\predSeq$. 
    A lot of papers focus on evaluating the calibration quality of specific language models (LMs) and tasks~\cite{wang-etal-2020-inference,kumar2019calibration}.
    Evaluation typically relies on variants of Expected Calibration Error (ECE)~\cite{tian-etal-2023-just,kumar2019calibration}.
    Oftentimes confidence scores from classification could be directly applied~\cite{stengel-eskin-van-durme-2023-calibrated} in order to evaluate whether an LM is over- or under-confident, especially for de facto classification tasks like sentiment analysis or multiple-choice QA.

\subsubsection{Confidence Estimation for LLMs} Confidence estimation in large language models (LLMs) refers to the task of quantifying how certain a model is about a specific generated output. In this subsection, we review three major families of approaches to confidence estimation in LLMs: 

\vspace{1mm}
\noindent$\bullet$~\textbf{UQ methods with Confidence Estimation.} As uncertainty and confidence are often intertwined, many approaches used in UQ have their counterpart in confidence estimation. 
For example, for black-box settings where the parameters of LLMs are unavailable, \cite{lin2023generating,zhang2024luq} computes a similarity matrix of sampled responses and derives confidence estimates for each generation via its degree or distance derived from the graph Laplacian, before using these scores to compute uncertainty.
For white-box settings where model parameters are available, researchers mostly compute the confidence from the output logits, either through normalizing \cref{eq:SL} with the length of $\predSeq$~~\cite{malinin2021uncertainty}, replacing the logit-sum or mean with weighted sum by attention values~\cite{lin2024contextualized} or by importance inferred from natural language inference (NLI) models~\cite{duan2024shifting}.
Such variants of sequence likelihood could then be fed for (entropy-style) uncertainty computation~\cite{kuhn2023semantic,lin2024contextualized}.

\vspace{1mm}
\noindent$\bullet$~\textbf{LLM-as-a-judge.} Another popular approach is asking the LM itself whether a particular free-form generation is correct~\cite{kadavath2022language}. However, this formulation also poses a restriction on the confidence estimation method, as it is essentially a scalar logit.
Thus, many extensions focus on applying calibration methods from classification to calibrate such self-evaluation.
The few exceptions include \cite{pmlr-v239-ren23a,kadavath2022language}, which converts samples from free-form generation into a \textit{multiple-choice} question (with generations being the options) and adds a "None of the above" option to elicit the confidence. 

\vspace{1mm}
\noindent$\bullet$~\textbf{Trainable Confidence Estimators.} Since we typically care about the LM's confidence in the ``semantic space'' due to semantic invariance, instead of manipulating logits, a popular approach is to perform additional training for confidence estimation. 
This could be done on the base LM (either fully~\cite{zhao2023calibrating,jiang-etal-2021-know,kapoor-etal-2024-calibration} or partially~\cite{liu2024litcab}) with a different loss, or using a separate model on the internal or external representations from the base LM~\cite{jagannatha-yu-2020-calibrating,ulmer-etal-2024-calibrating}.
On the other end of the spectrum, without any training, prompting could be used to elicit verbalized confidence values~\cite{tian-etal-2023-just}.
Finally, one could combine multiple confidence estimation methods and enjoy the benefit of ensembling~\cite{gao2024spuq}.

As with UQ evaluation (more in \cref{sec:eval}), the choice of correctness function has a profound impact on the conclusion of the experiments, especially for free-form generation tasks.
Popular choices include using (potentially larger) LLM as judges~\cite{liu2024litcab,lin2023generating,tian-etal-2023-just}, human annotations~\cite{pmlr-v239-ren23a}, or lexical similarities such as ROUGE~\cite{kuhn2023semantic,zhao2023calibrating}.
Recently, \citet{liu2025mcqa} proposes to evaluate free-form generation confidence measures with selected multiple-choice datasets as an efficient complement.
For longer generations, \citet{huang-etal-2024-calibrating} proposes to use ordinal (not binary) correctness values to capture the ambiguity in the quality of a generation. 
In a similar flavor, \cite{baan2022stop} studies the issues in the evaluation of calibration when there is intrinsic human disagreement on the label.

\begin{table*}
\centering
\resizebox{0.91\textwidth}{!}{
\begin{tabular}{lcccc}
\toprule
Method&Uncertainty Dimensions &Efficency Features&Access to Model&Confidence\\
\midrule
Input clarification ensembles~\cite{hou2024decomposing}&Input Uncertainty&Multi Rounds Generations&Black-box&No \\
ICL-Sample~\cite{ling2024uncertainty}&Input Uncertainty&Multi Rounds Generations&Black-box&No \\
SPUQ~\cite{gao2024spuq}&Input Uncertainty&Multi Rounds Generations + Additional Model &Black-box&No \\
\midrule
\midrule
UAG~\cite{yin2024reasoning}&Reasoning Uncertainty&Single Round Generation&White-box  &No\\
CoT-UQ~\cite{zhang2025cot}&Reasoning Uncertainty&Single Round Generation&White-box  &Yes\\
TouT~\cite{mo2024tree}&Reasoning Uncertainty&Multi Rounds Generations&Black-box &No\\
TopologyUQ~\cite{da2025understanding}&Reasoning Uncertainty&Multi Rounds Generations &Black-box &No\\

Stable Explanations Confidence~\cite{becker2024cycles} &Reasoning Uncertainty&Multi Rounds Generation& Black-box  &Yes\\

\midrule
\midrule
SAPLMA~\cite{azaria2023internal}&Parameter + Prediction Uncertainty&Fine-tuning&White-box  &Yes\\
Supervised estimation\cite{liu2024uncertainty}&Parameter + Prediction Uncertainty&Fine-tuning&White-box  &Yes\\
UaIT~\cite{liu2024can}&Parameter + Prediction Uncertainty&Fine-tuning&White-box&Yes\\
LoRA ensembles~\cite{balabanov2024uncertainty}&Parameter Uncertainty&Fine-tuning&White-box&Yes\\
BloB~\cite{wang2025blob}&Parameter Uncertainty&Fine-tuning&White-box&Yes\\
BLoRA~\cite{yangbayesian} & Parameter Uncertainty & Fine-tuning & White-box & Yes\\
\midrule
\midrule
Perplexity~\cite{mora2024uncertainty,margatina2023active}&Prediction Uncertainty&Single Round Generation&White-box&Yes\\
SAR~\cite{duan2024shifting}&Prediction Uncertainty&Single Round Generation&White-box  &Yes\\
P(True)~\cite{kadavath2022language}&Prediction Uncertainty&Single Round Generation&White-box  &Yes\\
Response improbability~\cite{fadeeva2024fact}&Prediction Uncertainty&Single Round Generation&White-box  &Yes\\
Average log probability~\cite{manakul2023selfcheckgpt}&Prediction Uncertainty&Single Round Generation&White-box  &Yes\\

Predictive Entropy~\cite{kadavath2022language}&Prediction Uncertainty&Multi Rounds Generations&White-box&Yes\\
Relative Mahalanobis distance~\cite{ren2022out}&Prediction Uncertainty&Multi Rounds Generations&White-box  &Yes\\
HUQ~\cite{vazhentsev2023huq}&Prediction Uncertainty&Multi Rounds Generations&White-box  &Yes\\

Conformal Prediction (CP)~\cite{quachconformal,kumar2023conformal}&Prediction Uncertainty&Multi Rounds Generations&White-box &No\\
ConU~\cite{wang2024conu}&Prediction Uncertainty&Multi Rounds Generations&White-box &No\\
Level-adaptive CP~\cite{cherian2025large}&Prediction Uncertainty&Multi Rounds Generations&White-box&No\\
LoFreeCP~\cite{su2024api}&Prediction Uncertainty&Multi Rounds Generations&Black-box&No\\
Ecc(J),Deg(J)~\cite{lin2023generating}&Prediction Uncertainty&Multi Rounds Generations&Black-box  &Yes\\
Eig(J)~\cite{lin2023generating}&Prediction Uncertainty&Multi Rounds Generations&Black-box  &No\\

Normal length predictive entropy~\cite{malinin2021uncertainty}&Prediction Uncertainty&Multi Rounds Generations +Additional Model&White-box  &Yes\\
Semantic Entropy~\cite{kuhn2023semantic}&Prediction Uncertainty&Multi Rounds Generations + Additional Model&White-box  &Yes\\
Kernel Language Entropy~\cite{nikitin2024kernel}&Prediction Uncertainty&Multi Rounds Generations + Additional Model&White-box  &Yes\\

Ecc(C),Ecc(E),Deg(C),Deg(E)~\cite{lin2023generating}&Prediction Uncertainty&Multi Rounds Generations + Additional Model&Black-box  &Yes\\
Eig(C),Eig(E)~\cite{lin2023generating}&Prediction Uncertainty&Multi Rounds Generations + Additional Model&Black-box  &No\\
MD-UQ~\cite{chen2025uncertainty} &Prediction Uncertainty&Multi Rounds Generations + Additional Model&Black-box  &No \\
D-UE~\cite{da2024llm} &Prediction Uncertainty&Multi Rounds Generations + Additional Model&Black-box  &Yes\\
\bottomrule
\end{tabular}
}
\caption{\small An overview of UQ methods discussed in this paper for different dimensions, efficiency, and model settings.}
\label{tab:Methods}
\vspace{-8mm}
\end{table*}

\vspace{1mm}
\noindent\textbf{Remarks.}
Existing literature sometimes uses the terms uncertainty and confidence interchangeably.
They do often seemingly coincide: When a model's prediction has low confidence, we naturally consider this as a high uncertainty case.
This, however, is treating $U(\xInput)=-\max_{\predSeq} \Conf(\xInput,\predSeq)$ as an uncertainty estimate.
In general, a model may exhibit high uncertainty over its output space but still express high confidence in a specific output. 
Conversely, a model could have low overall uncertainty but low confidence in a particular prediction.
While the ``low uncertainty low confidence case'' is relatively less interesting in classification or regression tasks due to MLE point prediction, this scenario is notably more common in NLG, as the output is typically \textit{randomly sampled}\footnote{In fact, even if the output is greedily generated, it might not have the highest confidence as measured by \cref{eq:SL}.} from the predictive distribution.
There are also applications that require one but not the other (e.g. conformal language modeling~\cite{quachconformal} or \textit{seletive generation}~\cite{cole2023selectively}). 
In the rest of this paper, we sometimes follow the language of the original papers and treat confidence estimates as uncertainty, but will clearly mark the methods that provide confidence estimates.
\section{UQ Methods for Different Dimensions}
\label{sec:uq-methods}

\subsection{Input Uncertainty}
As mentioned in \cref{sec:uncertainty_stage}, input uncertainty arises from the ambiguous or incomplete input to the LLMs. While there are works in the LLMs domain that try to benchmark or deal with ambiguity~\cite{guo2021abg-coqa,zamani2020generating,deng2023learning,chen2025abg}, they did not model the uncertainties induced by ambiguity.
Existing UQ methods that specifically consider input uncertainty focus on perturbing the input prompts of LLMs. For instance, \cite{hou2024decomposing} proposes an approach that generates multiple clarifications for a given prompt and ensembles the resulting generations by using mutual information to capture the disagreement among the predictions arising from different clarifications. Similarly, \cite{ling2024uncertainty} proposed ICL-Sample, which quantified the input uncertainty in the setting of in-context learning using different in-context samples. \cite{gao2024spuq} proposes SPUQ, which perturbs the input by techniques such as paraphrasing and dummy tokens to expose the model's sensitivity and capture uncertainty. Specifically, SPUQ quantified the input uncertainty by using a similarity metric such as BERTScore~\cite{zhangbertscore} to measure how consistent the responses are across different perturbations.
In general, there are only a few papers that consider input uncertainty. Since ambiguity is common and important in natural language, more effort is needed into input uncertainty and its application.

\subsection{Reasoning Uncertainty}
Reasoning is the process of drawing conclusions based on available information. As the LLMs have demonstrated remarkable performance on tasks involving reasoning, recent research has focused on using UQ in LLM reasoning and analyzing the internal reasoning process. For example, TopologyUQ~\cite{da2025understanding} introduces a formal method to extract and structure LLM explanations into graph representations, quantifying reasoning uncertainty by employing graph-edit distances and revealing redundancy through stable topology measures. Stable-Explanation Confidence~\cite{becker2024cycles} treats each possible model and its explanation pair as a test-time classifier to construct a posterior answer distribution that reflects overall reasoning confidence. CoT-UQ~\cite{zhang2025cot} integrates chain-of-thought reasoning into a response-level UQ framework, thereby leveraging the inherent multi-step reasoning capability of LLMs to further improve uncertainty assessment. Collectively, these approaches provide a robust and interpretable framework for enhancing LLM reasoning by quantifying uncertainty at local or global levels.

The quantified uncertainty could be used to guide the exploration of reasoning steps and improving the final performance in completing the tasks. In~\cite{mo2024tree}, they propose Tree of Uncertain Thoughts (TouT), which extend the Tree of Thoughts (ToT)~\cite{yao2023tree} framework by quantifying the uncertainties in intermediate reasoning steps with Monte Carlo Dropout and assigning uncertainty scores to important decision points. Similarly, ~\cite{yin2024reasoning} reduces the error accumulation in multi-step reasoning by monitoring the predicted probability of the next token at each generation step, 
dynamically retracting to more reliable states and incorporating certified reasoning clues when high uncertainty is detected. 
Their experimental results shows that integrating uncertainty enhances the precision of generated responses by integrating these local measures with global search techniques.

\subsection{Parameter Uncertainty}

Parameter uncertainty arises when an LLM lacks sufficient knowledge due to limitations in its training data or model parameters. It reflects the model’s uncertainty about its own predictions, which can be reduced with additional training or adaptation techniques.

Traditional UQ methods like Monte Carlo Dropout and Deep Ensembles have been widely used but are computationally infeasible for large-scale LLMs due to the need for multiple forward passes or model replicas. To address this, Bayesian Low-Rank Adaptation by Backpropagation (BLoB)~\cite{wang2025blob} and Bayesian Low-Rank Adaptation (BLoRA)~\cite{yangbayesian} incorporate Bayesian modeling into LoRA adapters, allowing uncertainty estimation through parameter distributions without a full-model ensemble. However, these methods still incur significant computational costs.

Finetuning-based approaches offer a more practical alternative. Techniques such as Supervised Uncertainty Estimation~\cite{liu2024uncertainty} train auxiliary models to predict the confidence of LLM outputs based on activation patterns and logit distributions. Similarly, Uncertainty-aware Instruction Tuning (UaIT)~\cite{liu2024can} modifies the fine-tuning process to explicitly train models to express uncertainty in their outputs. SAPLMA~\cite{azaria2023internal} refines probabilistic alignment techniques to dynamically adjust model uncertainty estimates, ensuring adaptability to different downstream tasks. Additionally, LoRA ensembles~\cite{balabanov2024uncertainty} provide an alternative to full-model ensembles by training multiple lightweight LoRA-adapted variants of an LLM instead of retraining the entire network. 


\subsection{Prediction Uncertainty}
Most off-the-shelf UQ methods focus on prediction uncertainty since it is the most straightforward way to estimate the uncertainty. Considering the number of generations and models when estimating uncertainties, existing methods for predicting uncertainty can be categorized into the following three categories.

\subsubsection{Single Round Generation}
Most single-round generation methods utilize the logit or hidden states during generation. With only one round of generation, these methods usually methods are usually efficient in estimating uncertainties.

\vspace{1mm}
\noindent$\bullet$~\textbf{Perplexity} is a measure of how well a probabilistic language model predicts a sequence of text~\cite{vaswani2017attention} while \citet{mora2024uncertainty}, \citet{margatina2023active} and \citet{manakul2023selfcheckgpt} utilize the perplexity as the uncertainty. In detail, using $w_i$ as the i-th token in the generation, perplexity is given by $
\text{Perplexity} = \exp \left( -\frac{1}{N} \sum_{i=1}^{N} \ln p(w_i) \right)$.
A higher perplexity means the model spreads its probability more broadly over possible words, indicating that it has a higher uncertainty.

\vspace{1mm}
\noindent$\bullet$~\textbf{Maximum Token Log-Probability.} Apart from the perplexity, Maximum token log-probability~\cite{manakul2023selfcheckgpt} measures the sentence’s likelihood by assessing the least likely token in the sentence. A higher $\text{Maximum}(p)$ indicates higher uncertainty of the whole generation. It is calculated by $
Max(p) =\max\limits_{i}(-\ln p(w_i))$.

\vspace{1mm}
\noindent$\bullet$~\textbf{Entropy} reflects how widely distributed a model’s predictions are for a given input, indicating the level of uncertainty in its outputs~\cite{kuhn2023semantic,kadavath2022language}. Entropy for the i-th token is provided by $
\mathcal{H}_{i} = - \sum_{\tilde{w} \in \mathcal{D}} p_i(\tilde{w}) \log p_i(\tilde{w})$.
Then it is possible to use the mean or maximum value of entropy as the final uncertainty~\cite{manakul2023selfcheckgpt}:
$
Avg(\mathcal{H}) = \frac{1}{N}\sum_{i=1}^N\mathcal{H}_i; Max(\mathcal{H}) = \max\limits_{i}(\mathcal{H}_i)
$.
Furthermore, Shifting Attention to Relevance (SAR)~\cite{duan2024shifting}, enhanced the performance of entropy by adjusting attention to more relevant tokens inside the sentence. In detail, SAR assigned weight for $\mathcal{H}_{i}$ and the weight $R(w_i, s, x)$ can be obtained by:
$R(w_i, s, x) = 1 - \left| g(x \cup s, x \cup s \setminus \{w_i\}) \right|,$
where $g$ is a function that measures the semantic similarity between two sentences, which can be estimated with NLI models~\cite{duan2024shifting}.

\vspace{1mm}
\noindent$\bullet$~\textbf{Response Improbability} \cite{fadeeva2024fact} uses response improbability, which computes the probability of a given sentence and subtracts the resulting value from one. In detail, response improbability is provided by $
    MP(s) = 1- \prod_{i=1} p_i(w_i).
$
If the sentence is certain (i.e., the product of token probabilities is high), $MP(s)$ will be low.

\vspace{1mm}
\noindent$\bullet$~\textbf{P(True)} \cite{kadavath2022language} measures the uncertainty of the claim by asking the LLM itself whether the generation is true or not. Specifically, P(True) is calculated~\footnote{The original name is P(IK), which stands for \textit{``I Know''}.}:
$
    \text{P(True)} = 1- p(y_1=\text{``True”}).
$
Note that here we are using $y_1$ as the first token instead of $w_1$ because $w_1$ represents the first token in the generation $s$ while $y_1$ represents the first token when asking LLM whether the generation $s$ is correct or not. 
P(True) requires running the LLM twice. However, it does not require multiple generations $s$. Therefore, we still classify this method as a single-round generation~\footnote{This could be considered an uncertainty estimate as the sequence to be evaluated is the prediction given the input.}.

\subsubsection{Multiple rounds generation} Multiple rounds generation methods estimate uncertainty by generating multiple predictions from the LLMs and analyzing their consistency, similarity, or variability. These approaches assume that if a model is confident, its outputs should be stable across different sampling conditions.

\vspace{1mm}
\noindent$\bullet$~\textbf{Token-Level Entropy.} Token-level entropy quantifies uncertainty in LLMs by analyzing the probability distribution of generated tokens across multiple samples. A confident model assigns high probability to a specific token, resulting in low entropy, while uncertain predictions distribute probability across multiple tokens, leading to higher entropy.

Multiple responses are generated for the same input to estimate token-level entropy, and the entropy of the token probability distribution is computed. For example, predictive entropy~\cite{kadavath2022language} can also be applied to multiple response settings and shows a better uncertainty quality based on the variability of multiple outputs. Similarly, SAR~\cite{duan2024shifting} could also be applied to multiple responses.~\cite{malinin2021uncertainty} extends with Monte Carlo-based approximations and focuses on how probability distributions evolve across tokens during autoregressive generation. There are two main approaches to get the final uncertainty: one averages entropy across multiple sampled outputs, and the other decomposes sequence-level uncertainty into token-level contributions using entropy approximation.

\vspace{1mm}
\noindent$\bullet$~\textbf{Conformal Prediction.} Conformal Prediction (CP)~\cite{shafer2008tutorial} is a statistical framework that provides formal coverage guarantees for uncertainty estimation in LLMs. Its distribution-free properties make it suitable for both black-box and white-box models.

In the black-box setting, where model internals are inaccessible, CP estimates uncertainty using response frequency, semantic similarity, or self-consistency. One study proposes a method tailored for API-only LLMs~\cite{su2024api}, using frequency-based sampling combined with normalized entropy and semantic similarity to define nonconformity scores. Another black-box CP method introduces a self-consistency-based uncertainty measure~\cite{wang2024conu}, which clusters sampled generations and selects a representative response to construct prediction sets with correctness guarantees, making it particularly effective for open-ended NLG tasks.

On the other hand, white-box CP methods use logits, internal activations, and calibration techniques for more refined uncertainty estimation. One study proposes Conformal Language Modeling~\cite{quachconformal}, which integrates CP with autoregressive text generation by dynamically calibrating a stopping rule to ensure at least one response in the generated set is statistically valid. Another work adapts CP for multiple-choice QA~\cite{kumar2023conformal}, using model confidence scores to calibrate prediction sets, ensuring coverage with minimal set size. A more advanced technique, conditional CP~\cite{cherian2025large}, dynamically adjusts coverage guarantees based on the difficulty of the input, optimizing prediction set size while maintaining reliability.

\vspace{1mm}
\noindent$\bullet$~\textbf{Consistency-Based Methods.} Consistency-based uncertainty estimation methods analyze the agreement between multiple generated responses from an LLM to determine uncertainty. The underlying assumption is that if the model is confident, its responses should be consistent, while high variability among responses suggests uncertainty. ~\cite{lin2023generating} measures the overlap between words through Jaccard similarity in different generations. This method evaluates the deviation from self-consistency, where a high Jaccard similarity across generations implies low uncertainty.

However, word-level similarity alone is insufficient, as different responses can convey the same meaning using different phrasing. Moreover, the generated response might include long reasoning steps that require detailed analysis~\cite{golovnevaroscoe}.
To address this problem, some methods incorporate external models to assess semantic similarity rather than relying solely on lexical overlap.

\subsubsection{Multiple Rounds Generation with External Models} Semantic-based uncertainty estimation methods expand multiple generation approaches by incorporating external models, such as Natural Language Inference (NLI) or pretrained language models, to evaluate the semantic relationships among generated responses beyond surface-level similarity.

\vspace{1mm}
\noindent$\bullet$~\textbf{Distribution-based entropy methods} quantify uncertainty by modeling the distribution of generated responses in a semantic space.
Semantic Entropy (SE)~\cite{kuhn2023semantic} refines uncertainty estimation by clustering generated responses based on semantic equivalence. This approach uses an NLI model to determine entailment relationships among responses, grouping them into meaning-preserving clusters. Instead of calculating entropy over individual responses, SE computes entropy over these clusters. 
Kernel Language Entropy (KLE)~\cite{nikitin2024kernel} takes a different approach by avoiding explicit clustering. Instead, it embeds the responses in a semantic space using a positive semidefinite kernel function. By computing von Neumann entropy over these response distributions, KLE provides an even more fine-grained measure of uncertainty that considers nuanced semantic variations.

\vspace{1mm}
\noindent$\bullet$~\textbf{Pairwise similarity methods} construct a pairwise semantic similarity matrix between responses and analyze its structural properties to estimate uncertainty.
Methods like~\cite{lin2023generating} use NLI models to score entailment and contradiction between every pair of generated outputs, forming a weighted similarity graph. A confident model yields semantically coherent responses with strong mutual agreement (high similarity), while inconsistent or ambiguous outputs lead to greater dispersion in the matrix. To quantify this dispersion, spectral graph metrics are applied: Eccentricity (Ecc) captures variability spread, Eigenvalue-based (Eig) measures assess global structure, and Degree (Deg) evaluates local consistency.
Recent works further extend this by modeling the response similarity graph as directed~\cite{da2024llm} or multi-dimensional~\cite{chen2025uncertainty}, allowing for richer representation of semantic asymmetry or latent factors in uncertainty.

\section{Evaluation of Uncertainty in LLMs}\label{sec:eval}

\subsection{Benchmark Datasets}

Datasets used in previous studies can be organized into several categories based on their focus.  An overall summary of the categorization of datasets and benchmarks for UQ is shown in \cref{tab:dataset}.

\vspace{1mm}
\noindent$\bullet$~\textbf{Reading comprehension} benchmarks include CoQA~\cite{reddy2019coqa} for conversational question answering tasks, RACE~\cite{lai2017race} for general reading comprehension, TriviaQA~\cite{joshi2017triviaqa} for fact-based questions, CosmosQA~\cite{huang2019cosmos} for contextual understanding, SQuAD~\cite{rajpurkar2016squad} for question answering on passages, and HotpotQA~\cite{yang2018hotpotqa} for multi-hop reasoning.

\vspace{1mm}
\noindent$\bullet$~\textbf{Reasoning and math} benchmarks include HotpotQA~\cite{yang2018hotpotqa} and StrategyQA~\cite{ Geva2021DidAU}, which test multi-hop reasoning, GSM8K~\cite{cobbe2021gsm8k} for solving math problems, and CalibratedMath~\cite{lin2022teaching}, designed to evaluate confidence expression in arithmetic. These benchmarks are helpful to evaluate the reasoning uncertainty.

\vspace{1mm}
\noindent$\bullet$~\textbf{Factuality} evaluation draws on datasets such as TruthfulQA~\cite{lin2021truthfulqa} for addressing common misconceptions, FEVER~\cite{thorne2018fever} for claim verification, HaluEval~\cite{li2023halueval} for detecting hallucinations, and an annotated FActScore~\cite{min2023factscore} dataset for evaluating the factuality of long-form text generated by LLMs.

\vspace{1mm}
\noindent$\bullet$~\textbf{General knowledge} benchmarks can be adapted for UQ to test the models' general knowledge, such as MMLU~\cite{hendrycks2020measuring} for a wide range of subjects, GPQA~\cite{rein2024gpqa} for multiple-choice questions in physical sciences, and HellaSwag~\cite{zellers2019hellaswag} for common-sense reasoning through sentence completion.
These benchmarks can be adapted for UQ because the tasks can be reduced to a classification problem, determining whether the model is confident or uncertain. The structured nature of these benchmarks allows for clear evaluation of the model’s confidence in its predictions.

\vspace{1mm}
\noindent$\bullet$~\textbf{Consistency and ambiguity} are two additional kind of benchmarks for UQ. Consistency benchmarks such as ParaRel~\cite{elazar2021measuring} tests semantic consistency across 328 paraphrases for 38 relations, and datasets like AmbigQA and AmbigInst, which feature inherent ambiguities~\cite{min2020ambigqa, hou2024decomposing,chen2025abg}.
Ambiguity datasets are useful in UQ evaluation because they introduce aleatoric uncertainty by highlighting cases where multiple plausible interpretations exist, helping to assess how well models distinguish between data-driven randomness and model-based uncertainty. These datasets enable a more precise decomposition of uncertainty into aleatoric and epistemic components, improving model reliability and interpretability.

\begin{table}[t]
\centering
\resizebox{!}{0.2\columnwidth}{
\begin{tabular}{p{3.3cm}p{5.6cm}}
\toprule
\textbf{Category} & \textbf{Benchmarks} \\
\midrule
Reading Comprehension & TriviaQA~\cite{joshi2017triviaqa}, CoQA~\cite{reddy2019coqa}, RACE~\cite{lai2017race}, CosmosQA~\cite{huang2019cosmos}, SQuAD~\cite{rajpurkar2016squad}, HotpotQA~\cite{yang2018hotpotqa} \\
\hline
Reasoning \& Math & StrategyQA~\cite{Geva2021DidAU}, HotpotQA~\cite{yang2018hotpotqa}, GSM8K~\cite{cobbe2021gsm8k}, CalibratedMath~\cite{lin2022teaching} \\
\hline
Factuality & TruthfulQA~\cite{lin2021truthfulqa}, FEVER~\cite{thorne2018fever}, HaluEval~\cite{li2023halueval},
FActScore~\cite{min2023factscore} \\
\hline
General Knowledge & MMLU~\cite{hendrycks2020measuring}, GPQA~\cite{rein2024gpqa}, HellaSwag~\cite{zellers2019hellaswag} \\
\hline
Consistency \& Ambiguity & ParaRel~\cite{elazar2021measuring}, AmbigQA~\cite{min2020ambigqa}, AmbigInst~\cite{hou2024decomposing},
Abg-SciQA~\cite{chen2025abg}
\\
\bottomrule
\end{tabular}}
\caption{\small Categorization of benchmarking datasets for UQ.}
\label{tab:dataset}
\vspace{-10mm}
\end{table}



Recently, there have been efforts to develop UQ benchmarks for dedicated sources of uncertainty or specific methods.
For example, MAQA~\cite{yang-etal-2025-maqa} is a dataset specifically designed to evaluate epistemic uncertainty in language models; LM-Polygraph~\cite{fadeeva2023lm} was later adopted as a comprehensive uncertainty benchmark~\cite{vashurin2024benchmarking}.
\cite{ye2025benchmarking} developed a benchmark for conformal prediction methods.
These contributions represent specialized datasets explicitly designed to assess UQ capabilities in LLMs, rather than adapting existing general-purpose benchmarks.

\subsection{Evaluation Metrics}
\label{sec:metrics}



UQ is often evaluated from binary classification tasks, with the rationale being that high uncertainty should correspond to low expected accuracy.
This is typically modeled by assigning a binary label to each response with a correctness function and using the uncertainty estimates to predict the label.
\texttt{AUROC} (Area Under the Receiver Operating Characteristic curve), which measures how effectively the uncertainty score separates correct from incorrect responses, is often used.
With values ranging from 0 to 1, higher \texttt{AUROC}s indicate better performance. 
Responses with confidence above the threshold are classified as predicted positives, while those below are treated as predicted negatives.
Many prior studies use \texttt{AUROC} to evaluate how well the uncertainty score discriminates correct from incorrect predictions~\cite{chen2023quantifying, kuhn2023semantic,xiong2024can,liu2024uncertainty,liu2025mcqa}.
Similarly, \texttt{AUPRC} (Area Under the Precision-Recall Curve) and \texttt{AUARC} (Area Under the Accuracy-Rejection Curve)~\cite{nadeem2009accuracy} also offer further insights into UQ. \texttt{AUPRC} measures how well the uncertainty score separates correct from incorrect responses~\cite{ling2024uncertainty}, while \texttt{AUARC} assesses how effectively the uncertainty measure aids in selecting accurate responses by determining which uncertain questions to reject~\cite{lin2024contextualized}. 

In the context of NLG where the correctness label is hard to obtain, researchers also compute heuristic-based fuzzy matching metrics such as \texttt{BLEU}~\cite{papineni2002bleu} and \texttt{ROUGE}~\cite{kuhn2023semantic} between the generated text and the reference output(s) to gauge the quality. 
However, these metrics often fail to capture semantic fidelity or factual correctness. Consequently, many researchers are increasingly turning to \texttt{LLM-as-a-judge} evaluations, wherein a large language model (e.g., GPT-4) is prompted to assess text quality or correctness. 
This approach can capture nuanced aspects like coherence, style, and factuality, but also introduces risks of bias and inconsistency. 
Human annotation, however, is expensive and is often limited to a small scale~\cite{zhang2024luq,kuhn2023semantic}. 


Apart from the binary classification framework, there are also multiple evaluation methods designed for the specific treatment of uncertainty, sometimes qualitative.
For example, focusing on decomposing aleatoric and epistemic uncertainty, \cite{hou2024decomposing} evaluates only the aleatoric part by using AmbigQA~\cite{min2020ambigqa}, as high ambiguity questions should incur higher aleatoric uncertainty (whereas math questions, for examples, might have lower).
The evaluation in \cite{giulianelli2023comes}, on the other hand, is a comparison between the variability of human production (generation) with that of the LM. 
With an emphasis on UQ for longer generations, \cite{zhang2024luq} compares the uncertainty estimate against FActScore~\cite{min2023factscore}, as the ``correctness'' of a long paragraph could be ill-defined or ambiguous.

\section{UQ Applications in LLMs}
LLMs are increasingly applied in diverse domains, offering flexibility and reasoning capabilities. However, UQ is crucial for ensuring their reliability, particularly in high-stakes applications. This section will introduce the applications that integrate the UQ of LLMs from some example domains.
Many other fields like energy management, operations research, etc., employ LLMs and would require such discussions on the need for UQ as well.

\vspace{1mm}
\noindent$\bullet$~\textbf{Robotics.} LLM-based robotic planning suffers from ambiguity and hallucinations, motivating the need for UQ in the planning loop. For example, closed-loop planners~\cite{zheng2024evaluating} employ an uncertainty-based failure detector to continuously assess and adjust plans in real-time, while non-parametric UQ methods~\cite{tsai2024efficient} use an efficient querying strategy to improve reliability. ~\cite{mullen2024lap} integrates action feasibility checks to align the LLM’s confidence with real-world constraints, improving success rates from approximately 70\% to 80\%. Similarly,~\cite{ong2024simple} dynamically adjusts thresholds for alternative paths in adaptive skill selection, achieving higher success rates. 
~\cite{liang2024introspective} develops an introspective planning framework with 
LLMs self-assess their uncertainty to enhance safety and human-robot collaboration.

\vspace{1mm}
\noindent$\bullet$~\textbf{Transportation.}
Preliminary research explores how LLMs can enhance transportation systems~\cite{da2024open,yao2025comal,da2023uncertainty}. For example, LLM inference has been used to bridge the sim-to-real gap in traffic signal control~\cite{da2023uncertainty,da2024prompt} and smooth mixed-autonomy traffic~\cite{yao2025comal}. However, both cases reveal the potential risk posed by hallucination. A few works have investigated the uncertainty measure while using the LLMs~\cite{de2023llm}, which tries to link the use of VLMs with deep probabilistic programming for UQ while conducting multimodal traffic accident forecasting tasks. 

\vspace{1mm}
\noindent$\bullet$~\textbf{Healthcare.}
In healthcare, LLMs and VLMs can be good references for diagnosis, but uncertainty is a critical dimension that should be considered together with the generation of more reliable treatment plans~\cite{savage2024large}. In~\cite{chen2024uncertainty}, it quantifies uncertainty in a white-box setting, and reveals that an effective reduction of model uncertainty can be achieved by using the proposed multi-tasking and ensemble methods in EHRs. However, as~\cite{wu2024uncertainty} benchmarks popular uncertain quantification methods with different model sizes on medical question-answering datasets, the challenge of UQ for medical applications is still severe. 

\vspace{1mm}
\section{Challenges and Future Directions}
While significant strides have been made in integrating uncertainty quantification into LLMs, several unaddressed challenges persist. This section will explore these unresolved issues, ranging from efficiency-performance trade-offs to cross-modal uncertainty, and outline promising avenues for future research, aiming to advance the reliability of LLMs in high-stakes applications.

\vspace{1mm}
\noindent$\bullet$~\textbf{Efficiency-Performance Trade-offs}. Multi-sample uncertainty methods incur prohibitive costs for trillion-parameter LLMs (\$12k per million queries~\cite{li2024llm}), yet yield marginal reliability gains ($\leq 0.02$ AUROC improvement~\cite{xiong2024efficient}). Hybrid approaches combining low-cost proxies (attention variance~\cite{heo2018uncertainty}, hidden state clustering~\cite{nikitin2024kernel}) could resolve this by achieving 90\% of maximal performance at 10\% computational cost. For example, precomputing uncertainty ``hotspots" during inference could trigger targeted multi-sampling only for high-risk outputs like medical diagnoses.


\vspace{1mm}
\noindent$\bullet$~\textbf{Interpretability Deficits}. Users struggle to distinguish whether uncertainty stems from ambiguous inputs, knowledge gaps, or decoding stochasticity. Modular architectures that decouple uncertainty estimation layers~\cite{huang2025latent,sensoy2018evidential} or employ causal tracing of transformer attention pathways \cite{wang2024grokking} could clarify uncertainty origins. For instance, perturbing model weights \cite{gal2016dropout} might reveal parametric uncertainty in low-resource languages, while input modules flag underspecified terms for clarification.

\vspace{1mm}
\noindent$\bullet$~\textbf{Cross-Modality Uncertainty}. Integrating vision, text, and sensor data introduces misaligned confidence estimates between modalities: LVLMs exhibit $2.4\times$ higher uncertainty in visual vs. textual components~\cite{zhang2024unveiling}, causing 63\% of errors in multi-modal QA~\cite{zhang2024vl}. Dynamic contrastive decoding and uncertainty-aware fusion protocols show promise\cite{huo2024self,suo2025octopus}, but require domain-specific adaptations (e.g., aligning radiology images with reports \cite{li2023unify,zhao2023radiology}). Future work must develop unified uncertainty embeddings to harmonize modality confidence scales and adversarial training against cross-modal backdoor attacks \cite{zhang2024badcm,liang2024revisiting}.

\vspace{1mm}
\noindent$\bullet$~\textbf{System-level Uncertainty in Agents and Reasoning.} 
As LLMs are increasingly deployed as autonomous agents or reasoning engines, the propagation and accumulation of uncertainty across steps becomes critical. Errors in early steps can lead to cascading failures, especially when the model expresses misplaced confidence. However, most existing UQ methods operate at one round of outputs from LLM, lacking mechanisms to capture uncertainty over multi-step reasoning chains or multi-action plans. As studies suggest that LLMs often fail to revise earlier decisions when presented with contradictory information~\cite{creswellselection}, there is a need for temporally-aware uncertainty tracking. Enhancing LLMs with structured memory or model-based planning, or leveraging graph-based representations to trace and revise uncertain steps~\cite{madaan2023self,yao2023tree}, could possibly provide more reliable behavior. 


\vspace{1mm}
\noindent$\bullet$~\textbf{UQ Evaluation.}
Evaluating the quality of UQ remains a fundamental challenge. 
While the binary classification metrics introduced in Section~\ref{sec:metrics} are widely used, they are not always suitable: many tasks, especially in NLG, cannot be easily reduced to binary correctness. 
Even for structured tasks like question answering, determining whether a free-form generation is correct can be nontrivial due to semantic variability and ambiguity. 
This issue becomes even more pronounced in open-ended tasks.
Moreover, LLM-as-a-judge evaluation approaches are themselves subject to systematic biases~\cite{panickssery2024llm,lin2021truthfulqa,zheng2023large}. 
In addition, common evaluation metrics such as \texttt{AUROC} and \texttt{AUARC} often fail to capture what might be considered ``meaningful" uncertainty. These metrics typically assess a model’s ability to distinguish between correct and incorrect outputs, but do not differentiate between confidently wrong responses and those accompanied by an appropriate level of uncertainty. 

\section{Conclusion}

In this survey, we offer a comprehensive overview of uncertainty quantification (UQ) in Large Language Models (LLMs). We first introduce the fundamental concepts relevant to both UQ and LLMs, highlighting the importance of reliability in high-stakes applications. Following this, we propose a detailed taxonomy for characterizing uncertainty dimensions in LLMs, including input, reasoning, parameter, and prediction uncertainty. We systematically introduce existing UQ methods using our novel taxonomy, reviewing their effectiveness across different uncertainty types. Ultimately, we identify and discuss some of the persistent challenges in UQ for LLMs, providing insightful directions for future research. The primary goal of this survey is to promote the integration of UQ techniques into LLM development, motivating both machine learning researchers and practitioners to participate in this rapidly advancing area.



\section*{ACKNOWLEDGMENTS}

The work was partially supported by NSF awards \#2421839. The views and conclusions contained in this paper are those of the authors and should not be interpreted as representing any funding agencies. We thank Amazon Research Awards and OpenAI for providing us with API credits under the Researcher Access program.


\bibliographystyle{ACM-Reference-Format}
\bibliography{main}


\begin{thebibliography}{141}


\ifx \showCODEN    \undefined \def \showCODEN     #1{\unskip}     \fi
\ifx \showISBNx    \undefined \def \showISBNx     #1{\unskip}     \fi
\ifx \showISBNxiii \undefined \def \showISBNxiii  #1{\unskip}     \fi
\ifx \showISSN     \undefined \def \showISSN      #1{\unskip}     \fi
\ifx \showLCCN     \undefined \def \showLCCN      #1{\unskip}     \fi
\ifx \shownote     \undefined \def \shownote      #1{#1}          \fi
\ifx \showarticletitle \undefined \def \showarticletitle #1{#1}   \fi
\ifx \showURL      \undefined \def \showURL       {\relax}        \fi
\providecommand\bibfield[2]{#2}
\providecommand\bibinfo[2]{#2}
\providecommand\natexlab[1]{#1}
\providecommand\showeprint[2][]{arXiv:#2}

\bibitem[Achiam et~al\mbox{.}(2023)]%
        {achiam2023gpt}
\bibfield{author}{\bibinfo{person}{Josh Achiam}, \bibinfo{person}{Steven Adler}, \bibinfo{person}{Sandhini Agarwal}, \bibinfo{person}{Lama Ahmad}, \bibinfo{person}{Ilge Akkaya}, \bibinfo{person}{Florencia~Leoni Aleman}, \bibinfo{person}{Diogo Almeida}, \bibinfo{person}{Janko Altenschmidt}, \bibinfo{person}{Sam Altman}, \bibinfo{person}{Shyamal Anadkat}, {et~al\mbox{.}}} \bibinfo{year}{2023}\natexlab{}.
\newblock \showarticletitle{Gpt-4 technical report}.
\newblock \bibinfo{journal}{\emph{arXiv preprint arXiv:2303.08774}} (\bibinfo{year}{2023}).
\newblock


\bibitem[Azaria and Mitchell(2023)]%
        {azaria2023internal}
\bibfield{author}{\bibinfo{person}{Amos Azaria} {and} \bibinfo{person}{Tom Mitchell}.} \bibinfo{year}{2023}\natexlab{}.
\newblock \showarticletitle{The Internal State of an {LLM} Knows When It`s Lying}.
\newblock  (\bibinfo{year}{2023}), \bibinfo{pages}{967--976}.
\newblock


\bibitem[Baan et~al\mbox{.}(2022)]%
        {baan2022stop}
\bibfield{author}{\bibinfo{person}{Joris Baan}, \bibinfo{person}{Wilker Aziz}, \bibinfo{person}{Barbara Plank}, {and} \bibinfo{person}{Raquel Fern{\'a}ndez}.} \bibinfo{year}{2022}\natexlab{}.
\newblock \showarticletitle{Stop Measuring Calibration When Humans Disagree}. In \bibinfo{booktitle}{\emph{Proceedings of the 2022 Conference on Empirical Methods in Natural Language Processing}}. \bibinfo{pages}{1892--1915}.
\newblock


\bibitem[Balabanov and Linander(2024)]%
        {balabanov2024uncertainty}
\bibfield{author}{\bibinfo{person}{Oleksandr Balabanov} {and} \bibinfo{person}{Hampus Linander}.} \bibinfo{year}{2024}\natexlab{}.
\newblock \showarticletitle{Uncertainty quantification in fine-tuned LLMs using LoRA ensembles}.
\newblock \bibinfo{journal}{\emph{arXiv preprint arXiv:2402.12264}} (\bibinfo{year}{2024}).
\newblock


\bibitem[Becker and Soatto(2024)]%
        {becker2024cycles}
\bibfield{author}{\bibinfo{person}{Evan Becker} {and} \bibinfo{person}{Stefano Soatto}.} \bibinfo{year}{2024}\natexlab{}.
\newblock \showarticletitle{Cycles of thought: Measuring llm confidence through stable explanations}.
\newblock \bibinfo{journal}{\emph{arXiv preprint arXiv:2406.03441}} (\bibinfo{year}{2024}).
\newblock


\bibitem[Chen and Mueller(2024)]%
        {chen2023quantifying}
\bibfield{author}{\bibinfo{person}{Jiuhai Chen} {and} \bibinfo{person}{Jonas Mueller}.} \bibinfo{year}{2024}\natexlab{}.
\newblock \showarticletitle{Quantifying Uncertainty in Answers from any Language Model and Enhancing their Trustworthiness}. In \bibinfo{booktitle}{\emph{Proceedings of the 62nd Annual Meeting of the Association for Computational Linguistics}}. \bibinfo{pages}{5186--5200}.
\newblock


\bibitem[Chen et~al\mbox{.}(2025a)]%
        {chen2025abg}
\bibfield{author}{\bibinfo{person}{Tiejin Chen}, \bibinfo{person}{Kuan-Ru Liou}, \bibinfo{person}{Mithun Shivakoti}, \bibinfo{person}{Aaryan Gaur}, \bibinfo{person}{Pragya Kumari}, \bibinfo{person}{Meiqi Guo}, {and} \bibinfo{person}{Hua Wei}.} \bibinfo{year}{2025}\natexlab{a}.
\newblock \showarticletitle{Abg-SciQA: A dataset for Understanding and Resolving Ambiguity in Scientific Questions}. In \bibinfo{booktitle}{\emph{ICLR 2025 Workshop on Navigating and Addressing Data Problems for Foundation Models}}.
\newblock


\bibitem[Chen et~al\mbox{.}(2025b)]%
        {chen2025uncertainty}
\bibfield{author}{\bibinfo{person}{Tiejin Chen}, \bibinfo{person}{Xiaoou Liu}, \bibinfo{person}{Longchao Da}, \bibinfo{person}{Jia Chen}, \bibinfo{person}{Vagelis Papalexakis}, {and} \bibinfo{person}{Hua Wei}.} \bibinfo{year}{2025}\natexlab{b}.
\newblock \showarticletitle{Uncertainty Quantification of Large Language Models through Multi-Dimensional Responses}.
\newblock \bibinfo{journal}{\emph{arXiv preprint arXiv:2502.16820}} (\bibinfo{year}{2025}).
\newblock


\bibitem[Chen et~al\mbox{.}(2024)]%
        {chen2024uncertainty}
\bibfield{author}{\bibinfo{person}{Zizhang Chen}, \bibinfo{person}{Peizhao Li}, \bibinfo{person}{Xiaomeng Dong}, {and} \bibinfo{person}{Pengyu Hong}.} \bibinfo{year}{2024}\natexlab{}.
\newblock \showarticletitle{Uncertainty Quantification for Clinical Outcome Predictions with (Large) Language Models}.
\newblock \bibinfo{journal}{\emph{arXiv preprint arXiv:2411.03497}} (\bibinfo{year}{2024}).
\newblock


\bibitem[Cheong et~al\mbox{.}(2024)]%
        {cheong2024not}
\bibfield{author}{\bibinfo{person}{Inyoung Cheong}, \bibinfo{person}{King Xia}, \bibinfo{person}{KJ~Kevin Feng}, \bibinfo{person}{Quan~Ze Chen}, {and} \bibinfo{person}{Amy~X Zhang}.} \bibinfo{year}{2024}\natexlab{}.
\newblock \showarticletitle{(A) I am not A lawyer, but...: engaging legal experts towards responsible LLM policies for legal advice}. In \bibinfo{booktitle}{\emph{Proceedings of the 2024 ACM Conference on Fairness, Accountability, and Transparency}}. \bibinfo{pages}{2454--2469}.
\newblock


\bibitem[Cherian et~al\mbox{.}(2025)]%
        {cherian2025large}
\bibfield{author}{\bibinfo{person}{John Cherian}, \bibinfo{person}{Isaac Gibbs}, {and} \bibinfo{person}{Emmanuel Candes}.} \bibinfo{year}{2025}\natexlab{}.
\newblock \showarticletitle{Large language model validity via enhanced conformal prediction methods}.
\newblock \bibinfo{journal}{\emph{Advances in Neural Information Processing Systems}}  \bibinfo{volume}{37} (\bibinfo{year}{2025}), \bibinfo{pages}{114812--114842}.
\newblock


\bibitem[Cobbe et~al\mbox{.}(2021)]%
        {cobbe2021gsm8k}
\bibfield{author}{\bibinfo{person}{Karl Cobbe}, \bibinfo{person}{Vineet Kosaraju}, \bibinfo{person}{Mohammad Bavarian}, \bibinfo{person}{Mark Chen}, \bibinfo{person}{Heewoo Jun}, \bibinfo{person}{Lukasz Kaiser}, \bibinfo{person}{Matthias Plappert}, \bibinfo{person}{Jerry Tworek}, \bibinfo{person}{Jacob Hilton}, \bibinfo{person}{Reiichiro Nakano}, \bibinfo{person}{Christopher Hesse}, {and} \bibinfo{person}{John Schulman}.} \bibinfo{year}{2021}\natexlab{}.
\newblock \showarticletitle{Training Verifiers to Solve Math Word Problems}.
\newblock \bibinfo{journal}{\emph{arXiv preprint arXiv:2110.14168}} (\bibinfo{year}{2021}).
\newblock


\bibitem[Cole et~al\mbox{.}(2023)]%
        {cole2023selectively}
\bibfield{author}{\bibinfo{person}{Jeremy Cole}, \bibinfo{person}{Michael Zhang}, \bibinfo{person}{Dan Gillick}, \bibinfo{person}{Julian Eisenschlos}, \bibinfo{person}{Bhuwan Dhingra}, {and} \bibinfo{person}{Jacob Eisenstein}.} \bibinfo{year}{2023}\natexlab{}.
\newblock \showarticletitle{Selectively Answering Ambiguous Questions}. In \bibinfo{booktitle}{\emph{Proceedings of the 2023 Conference on Empirical Methods in Natural Language Processing}}. \bibinfo{pages}{530--543}.
\newblock


\bibitem[Creswell et~al\mbox{.}({[n.\,d.]})]%
        {creswellselection}
\bibfield{author}{\bibinfo{person}{Antonia Creswell}, \bibinfo{person}{Murray Shanahan}, {and} \bibinfo{person}{Irina Higgins}.} \bibinfo{year}{[n.\,d.]}\natexlab{}.
\newblock \showarticletitle{Selection-Inference: Exploiting Large Language Models for Interpretable Logical Reasoning}. In \bibinfo{booktitle}{\emph{The Eleventh International Conference on Learning Representations}}.
\newblock


\bibitem[Da et~al\mbox{.}(2024a)]%
        {da2024llm}
\bibfield{author}{\bibinfo{person}{Longchao Da}, \bibinfo{person}{Tiejin Chen}, \bibinfo{person}{Lu Cheng}, {and} \bibinfo{person}{Hua Wei}.} \bibinfo{year}{2024}\natexlab{a}.
\newblock \showarticletitle{Llm uncertainty quantification through directional entailment graph and claim level response augmentation}.
\newblock \bibinfo{journal}{\emph{arXiv preprint arXiv:2407.00994}} (\bibinfo{year}{2024}).
\newblock


\bibitem[Da et~al\mbox{.}(2024b)]%
        {da2024prompt}
\bibfield{author}{\bibinfo{person}{Longchao Da}, \bibinfo{person}{Minquan Gao}, \bibinfo{person}{Hao Mei}, {and} \bibinfo{person}{Hua Wei}.} \bibinfo{year}{2024}\natexlab{b}.
\newblock \showarticletitle{Prompt to transfer: Sim-to-real transfer for traffic signal control with prompt learning}. In \bibinfo{booktitle}{\emph{Proceedings of the AAAI Conference on Artificial Intelligence}}, Vol.~\bibinfo{volume}{38}. \bibinfo{pages}{82--90}.
\newblock


\bibitem[Da et~al\mbox{.}(2024c)]%
        {da2024open}
\bibfield{author}{\bibinfo{person}{Longchao Da}, \bibinfo{person}{Kuanru Liou}, \bibinfo{person}{Tiejin Chen}, \bibinfo{person}{Xuesong Zhou}, \bibinfo{person}{Xiangyong Luo}, \bibinfo{person}{Yezhou Yang}, {and} \bibinfo{person}{Hua Wei}.} \bibinfo{year}{2024}\natexlab{c}.
\newblock \showarticletitle{Open-ti: Open traffic intelligence with augmented language model}.
\newblock \bibinfo{journal}{\emph{International Journal of Machine Learning and Cybernetics}} \bibinfo{volume}{15}, \bibinfo{number}{10} (\bibinfo{year}{2024}), \bibinfo{pages}{4761--4786}.
\newblock


\bibitem[Da et~al\mbox{.}(2025)]%
        {da2025understanding}
\bibfield{author}{\bibinfo{person}{Longchao Da}, \bibinfo{person}{Xiaoou Liu}, \bibinfo{person}{Jiaxin Dai}, \bibinfo{person}{Lu Cheng}, \bibinfo{person}{Yaqing Wang}, {and} \bibinfo{person}{Hua Wei}.} \bibinfo{year}{2025}\natexlab{}.
\newblock \showarticletitle{Understanding the Uncertainty of LLM Explanations: A Perspective Based on Reasoning Topology}.
\newblock \bibinfo{journal}{\emph{arXiv preprint arXiv:2502.17026}} (\bibinfo{year}{2025}).
\newblock


\bibitem[Da et~al\mbox{.}(2023)]%
        {da2023uncertainty}
\bibfield{author}{\bibinfo{person}{Longchao Da}, \bibinfo{person}{Hao Mei}, \bibinfo{person}{Romir Sharma}, {and} \bibinfo{person}{Hua Wei}.} \bibinfo{year}{2023}\natexlab{}.
\newblock \showarticletitle{Uncertainty-aware grounded action transformation towards sim-to-real transfer for traffic signal control}. In \bibinfo{booktitle}{\emph{2023 62nd IEEE Conference on Decision and Control (CDC)}}. \bibinfo{pages}{1124--1129}.
\newblock


\bibitem[Da et~al\mbox{.}(2024d)]%
        {da2024segment}
\bibfield{author}{\bibinfo{person}{Longchao Da}, \bibinfo{person}{Rui Wang}, \bibinfo{person}{Xiaojian Xu}, \bibinfo{person}{Parminder Bhatia}, \bibinfo{person}{Taha Kass-Hout}, \bibinfo{person}{Hua Wei}, {and} \bibinfo{person}{Cao Xiao}.} \bibinfo{year}{2024}\natexlab{d}.
\newblock \showarticletitle{Segment as You Wish--Free-Form Language-Based Segmentation for Medical Images}.
\newblock \bibinfo{journal}{\emph{arXiv preprint arXiv:2410.12831}} (\bibinfo{year}{2024}).
\newblock


\bibitem[de~Zarz{\`a} et~al\mbox{.}(2023)]%
        {de2023llm}
\bibfield{author}{\bibinfo{person}{Irene de Zarz{\`a}}, \bibinfo{person}{Joachim de Curt{\`o}}, \bibinfo{person}{Gemma Roig}, {and} \bibinfo{person}{Carlos~T Calafate}.} \bibinfo{year}{2023}\natexlab{}.
\newblock \showarticletitle{LLM multimodal traffic accident forecasting}.
\newblock \bibinfo{journal}{\emph{Sensors}} \bibinfo{volume}{23}, \bibinfo{number}{22} (\bibinfo{year}{2023}), \bibinfo{pages}{9225}.
\newblock


\bibitem[Deng et~al\mbox{.}(2023)]%
        {deng2023learning}
\bibfield{author}{\bibinfo{person}{Yang Deng}, \bibinfo{person}{Shuaiyi Li}, {and} \bibinfo{person}{Wai Lam}.} \bibinfo{year}{2023}\natexlab{}.
\newblock \showarticletitle{Learning to ask clarification questions with spatial reasoning}. In \bibinfo{booktitle}{\emph{Proceedings of the 46th International ACM SIGIR Conference on Research and Development in Information Retrieval}}. \bibinfo{pages}{2113--2117}.
\newblock


\bibitem[Der~Kiureghian and Ditlevsen(2009)]%
        {der2009aleatory}
\bibfield{author}{\bibinfo{person}{Armen Der~Kiureghian} {and} \bibinfo{person}{Ove Ditlevsen}.} \bibinfo{year}{2009}\natexlab{}.
\newblock \showarticletitle{Aleatory or epistemic? Does it matter?}
\newblock \bibinfo{journal}{\emph{Structural safety}} \bibinfo{volume}{31}, \bibinfo{number}{2} (\bibinfo{year}{2009}), \bibinfo{pages}{105--112}.
\newblock


\bibitem[Duan et~al\mbox{.}(2024)]%
        {duan2024shifting}
\bibfield{author}{\bibinfo{person}{Jinhao Duan}, \bibinfo{person}{Hao Cheng}, \bibinfo{person}{Shiqi Wang}, \bibinfo{person}{Alex Zavalny}, \bibinfo{person}{Chenan Wang}, \bibinfo{person}{Renjing Xu}, \bibinfo{person}{Bhavya Kailkhura}, {and} \bibinfo{person}{Kaidi Xu}.} \bibinfo{year}{2024}\natexlab{}.
\newblock \showarticletitle{Shifting Attention to Relevance: Towards the Predictive Uncertainty Quantification of Free-Form Large Language Models}. In \bibinfo{booktitle}{\emph{Proceedings of the 62nd Annual Meeting of the Association for Computational Linguistics (Volume 1: Long Papers)}}. \bibinfo{pages}{5050--5063}.
\newblock


\bibitem[Elazar et~al\mbox{.}(2021)]%
        {elazar2021measuring}
\bibfield{author}{\bibinfo{person}{Yanai Elazar}, \bibinfo{person}{Nora Kassner}, \bibinfo{person}{Shauli Ravfogel}, \bibinfo{person}{Abhilasha Ravichander}, \bibinfo{person}{Eduard Hovy}, \bibinfo{person}{Hinrich Sch{\"u}tze}, {and} \bibinfo{person}{Yoav Goldberg}.} \bibinfo{year}{2021}\natexlab{}.
\newblock \showarticletitle{Measuring and improving consistency in pretrained language models}.
\newblock \bibinfo{journal}{\emph{Transactions of the Association for Computational Linguistics}}  \bibinfo{volume}{9} (\bibinfo{year}{2021}), \bibinfo{pages}{1012--1031}.
\newblock


\bibitem[Fadeeva et~al\mbox{.}(2024)]%
        {fadeeva2024fact}
\bibfield{author}{\bibinfo{person}{Ekaterina Fadeeva}, \bibinfo{person}{Aleksandr Rubashevskii}, \bibinfo{person}{Artem Shelmanov}, \bibinfo{person}{Sergey Petrakov}, \bibinfo{person}{Haonan Li}, \bibinfo{person}{Hamdy Mubarak}, \bibinfo{person}{Evgenii Tsymbalov}, \bibinfo{person}{Gleb Kuzmin}, \bibinfo{person}{Alexander Panchenko}, \bibinfo{person}{Timothy Baldwin}, {et~al\mbox{.}}} \bibinfo{year}{2024}\natexlab{}.
\newblock \showarticletitle{Fact-Checking the Output of Large Language Models via Token-Level Uncertainty Quantification}. In \bibinfo{booktitle}{\emph{Findings of the Association for Computational Linguistics ACL 2024}}. \bibinfo{pages}{9367--9385}.
\newblock


\bibitem[Fadeeva et~al\mbox{.}(2023)]%
        {fadeeva2023lm}
\bibfield{author}{\bibinfo{person}{Ekaterina Fadeeva}, \bibinfo{person}{Roman Vashurin}, \bibinfo{person}{Akim Tsvigun}, \bibinfo{person}{Artem Vazhentsev}, \bibinfo{person}{Sergey Petrakov}, \bibinfo{person}{Kirill Fedyanin}, \bibinfo{person}{Daniil Vasilev}, \bibinfo{person}{Elizaveta Goncharova}, \bibinfo{person}{Alexander Panchenko}, \bibinfo{person}{Maxim Panov}, {et~al\mbox{.}}} \bibinfo{year}{2023}\natexlab{}.
\newblock \showarticletitle{LM-Polygraph: Uncertainty Estimation for Language Models}. In \bibinfo{booktitle}{\emph{Proceedings of the 2023 Conference on Empirical Methods in Natural Language Processing: System Demonstrations}}. \bibinfo{pages}{446--461}.
\newblock


\bibitem[Gal and Ghahramani(2016)]%
        {gal2016dropout}
\bibfield{author}{\bibinfo{person}{Yarin Gal} {and} \bibinfo{person}{Zoubin Ghahramani}.} \bibinfo{year}{2016}\natexlab{}.
\newblock \showarticletitle{Dropout as a bayesian approximation: Representing model uncertainty in deep learning}. In \bibinfo{booktitle}{\emph{international conference on machine learning}}. PMLR, \bibinfo{pages}{1050--1059}.
\newblock


\bibitem[Gao et~al\mbox{.}(2024)]%
        {gao2024spuq}
\bibfield{author}{\bibinfo{person}{Xiang Gao}, \bibinfo{person}{Jiaxin Zhang}, \bibinfo{person}{Lalla Mouatadid}, {and} \bibinfo{person}{Kamalika Das}.} \bibinfo{year}{2024}\natexlab{}.
\newblock \showarticletitle{SPUQ: Perturbation-Based Uncertainty Quantification for Large Language Models}. In \bibinfo{booktitle}{\emph{Proceedings of the 18th Conference of the European Chapter of the Association for Computational Linguistics (Volume 1: Long Papers)}}. \bibinfo{pages}{2336--2346}.
\newblock


\bibitem[Geva et~al\mbox{.}(2021)]%
        {Geva2021DidAU}
\bibfield{author}{\bibinfo{person}{Mor Geva}, \bibinfo{person}{Daniel Khashabi}, \bibinfo{person}{Elad Segal}, \bibinfo{person}{Tushar Khot}, \bibinfo{person}{Dan Roth}, {and} \bibinfo{person}{Jonathan Berant}.} \bibinfo{year}{2021}\natexlab{}.
\newblock \showarticletitle{Did Aristotle Use a Laptop? A Question Answering Benchmark with Implicit Reasoning Strategies}.
\newblock \bibinfo{journal}{\emph{Transactions of the Association for Computational Linguistics}} (\bibinfo{year}{2021}), \bibinfo{pages}{346--361}.
\newblock


\bibitem[Giulianelli et~al\mbox{.}(2023)]%
        {giulianelli2023comes}
\bibfield{author}{\bibinfo{person}{Mario Giulianelli}, \bibinfo{person}{Joris Baan}, \bibinfo{person}{Wilker Aziz}, \bibinfo{person}{Raquel Fern{\'a}ndez}, {and} \bibinfo{person}{Barbara Plank}.} \bibinfo{year}{2023}\natexlab{}.
\newblock \showarticletitle{What Comes Next? Evaluating Uncertainty in Neural Text Generators Against Human Production Variability}. In \bibinfo{booktitle}{\emph{Proceedings of the 2023 Conference on Empirical Methods in Natural Language Processing}}. \bibinfo{pages}{14349--14371}.
\newblock


\bibitem[Golovneva et~al\mbox{.}({[n.\,d.]})]%
        {golovnevaroscoe}
\bibfield{author}{\bibinfo{person}{Olga Golovneva}, \bibinfo{person}{Moya~Peng Chen}, \bibinfo{person}{Spencer Poff}, \bibinfo{person}{Martin Corredor}, \bibinfo{person}{Luke Zettlemoyer}, \bibinfo{person}{Maryam Fazel-Zarandi}, {and} \bibinfo{person}{Asli Celikyilmaz}.} \bibinfo{year}{[n.\,d.]}\natexlab{}.
\newblock \showarticletitle{ROSCOE: A Suite of Metrics for Scoring Step-by-Step Reasoning}. In \bibinfo{booktitle}{\emph{The Eleventh International Conference on Learning Representations}}.
\newblock


\bibitem[Guo et~al\mbox{.}(2021)]%
        {guo2021abg-coqa}
\bibfield{author}{\bibinfo{person}{Meiqi Guo}, \bibinfo{person}{Mingda Zhang}, \bibinfo{person}{Siva Reddy}, {and} \bibinfo{person}{Malihe Alikhani}.} \bibinfo{year}{2021}\natexlab{}.
\newblock \showarticletitle{Abg-coqa: Clarifying ambiguity in conversational question answering}.
\newblock \bibinfo{journal}{\emph{3rd Conference on Automated Knowledge Base Construction}} (\bibinfo{year}{2021}).
\newblock


\bibitem[Hendrycks et~al\mbox{.}(2020)]%
        {hendrycks2020measuring}
\bibfield{author}{\bibinfo{person}{Dan Hendrycks}, \bibinfo{person}{Collin Burns}, \bibinfo{person}{Steven Basart}, \bibinfo{person}{Andy Zou}, \bibinfo{person}{Mantas Mazeika}, \bibinfo{person}{Dawn Song}, {and} \bibinfo{person}{Jacob Steinhardt}.} \bibinfo{year}{2020}\natexlab{}.
\newblock \showarticletitle{Measuring massive multitask language understanding}.
\newblock \bibinfo{journal}{\emph{arXiv preprint arXiv:2009.03300}} (\bibinfo{year}{2020}).
\newblock


\bibitem[Heo et~al\mbox{.}(2018)]%
        {heo2018uncertainty}
\bibfield{author}{\bibinfo{person}{Jay Heo}, \bibinfo{person}{Hae~Beom Lee}, \bibinfo{person}{Saehoon Kim}, \bibinfo{person}{Juho Lee}, \bibinfo{person}{Kwang~Joon Kim}, \bibinfo{person}{Eunho Yang}, {and} \bibinfo{person}{Sung~Ju Hwang}.} \bibinfo{year}{2018}\natexlab{}.
\newblock \showarticletitle{Uncertainty-aware attention for reliable interpretation and prediction}.
\newblock \bibinfo{journal}{\emph{Advances in neural information processing systems}}  \bibinfo{volume}{31} (\bibinfo{year}{2018}).
\newblock


\bibitem[Hou et~al\mbox{.}(2024)]%
        {hou2024decomposing}
\bibfield{author}{\bibinfo{person}{Bairu Hou}, \bibinfo{person}{Yujian Liu}, \bibinfo{person}{Kaizhi Qian}, \bibinfo{person}{Jacob Andreas}, \bibinfo{person}{Shiyu Chang}, {and} \bibinfo{person}{Yang Zhang}.} \bibinfo{year}{2024}\natexlab{}.
\newblock \showarticletitle{Decomposing Uncertainty for Large Language Models through Input Clarification Ensembling}. In \bibinfo{booktitle}{\emph{Forty-first International Conference on Machine Learning}}.
\newblock


\bibitem[Huang et~al\mbox{.}(2024b)]%
        {huang2024survey}
\bibfield{author}{\bibinfo{person}{Hsiu-Yuan Huang}, \bibinfo{person}{Yutong Yang}, \bibinfo{person}{Zhaoxi Zhang}, \bibinfo{person}{Sanwoo Lee}, {and} \bibinfo{person}{Yunfang Wu}.} \bibinfo{year}{2024}\natexlab{b}.
\newblock \showarticletitle{A survey of uncertainty estimation in llms: Theory meets practice}.
\newblock \bibinfo{journal}{\emph{arXiv preprint arXiv:2410.15326}} (\bibinfo{year}{2024}).
\newblock


\bibitem[Huang et~al\mbox{.}(2025b)]%
        {huang2025latent}
\bibfield{author}{\bibinfo{person}{Jingwang Huang}, \bibinfo{person}{Jiang Zhong}, \bibinfo{person}{Qin Lei}, \bibinfo{person}{Jinpeng Gao}, \bibinfo{person}{Yuming Yang}, \bibinfo{person}{Sirui Wang}, \bibinfo{person}{Peiguang Li}, {and} \bibinfo{person}{Kaiwen Wei}.} \bibinfo{year}{2025}\natexlab{b}.
\newblock \showarticletitle{Latent Distribution Decoupling: A Probabilistic Framework for Uncertainty-Aware Multimodal Emotion Recognition}.
\newblock \bibinfo{journal}{\emph{arXiv preprint arXiv:2502.13954}} (\bibinfo{year}{2025}).
\newblock


\bibitem[Huang et~al\mbox{.}(2019)]%
        {huang2019cosmos}
\bibfield{author}{\bibinfo{person}{Lifu Huang}, \bibinfo{person}{Ronan~Le Bras}, \bibinfo{person}{Chandra Bhagavatula}, {and} \bibinfo{person}{Yejin Choi}.} \bibinfo{year}{2019}\natexlab{}.
\newblock \showarticletitle{Cosmos QA: Machine reading comprehension with contextual commonsense reasoning}.
\newblock \bibinfo{journal}{\emph{arXiv preprint arXiv:1909.00277}} (\bibinfo{year}{2019}).
\newblock


\bibitem[Huang et~al\mbox{.}(2025a)]%
        {huang2025survey}
\bibfield{author}{\bibinfo{person}{Lei Huang}, \bibinfo{person}{Weijiang Yu}, \bibinfo{person}{Weitao Ma}, \bibinfo{person}{Weihong Zhong}, \bibinfo{person}{Zhangyin Feng}, \bibinfo{person}{Haotian Wang}, \bibinfo{person}{Qianglong Chen}, \bibinfo{person}{Weihua Peng}, \bibinfo{person}{Xiaocheng Feng}, \bibinfo{person}{Bing Qin}, {et~al\mbox{.}}} \bibinfo{year}{2025}\natexlab{a}.
\newblock \showarticletitle{A survey on hallucination in large language models: Principles, taxonomy, challenges, and open questions}.
\newblock \bibinfo{journal}{\emph{ACM Transactions on Information Systems}} \bibinfo{volume}{43}, \bibinfo{number}{2} (\bibinfo{year}{2025}), \bibinfo{pages}{1--55}.
\newblock


\bibitem[Huang et~al\mbox{.}(2024a)]%
        {huang-etal-2024-calibrating}
\bibfield{author}{\bibinfo{person}{Yukun Huang}, \bibinfo{person}{Yixin Liu}, \bibinfo{person}{Raghuveer Thirukovalluru}, \bibinfo{person}{Arman Cohan}, {and} \bibinfo{person}{Bhuwan Dhingra}.} \bibinfo{year}{2024}\natexlab{a}.
\newblock \showarticletitle{Calibrating Long-form Generations From Large Language Models}. In \bibinfo{booktitle}{\emph{Findings of the Association for Computational Linguistics: EMNLP 2024}}. \bibinfo{pages}{13441--13460}.
\newblock


\bibitem[H{\"u}llermeier and Waegeman(2021)]%
        {hullermeier2021aleatoric}
\bibfield{author}{\bibinfo{person}{Eyke H{\"u}llermeier} {and} \bibinfo{person}{Willem Waegeman}.} \bibinfo{year}{2021}\natexlab{}.
\newblock \showarticletitle{Aleatoric and epistemic uncertainty in machine learning: An introduction to concepts and methods}.
\newblock \bibinfo{journal}{\emph{Machine learning}} \bibinfo{volume}{110}, \bibinfo{number}{3} (\bibinfo{year}{2021}), \bibinfo{pages}{457--506}.
\newblock


\bibitem[Huo et~al\mbox{.}(2024)]%
        {huo2024self}
\bibfield{author}{\bibinfo{person}{Fushuo Huo}, \bibinfo{person}{Wenchao Xu}, \bibinfo{person}{Zhong Zhang}, \bibinfo{person}{Haozhao Wang}, \bibinfo{person}{Zhicheng Chen}, {and} \bibinfo{person}{Peilin Zhao}.} \bibinfo{year}{2024}\natexlab{}.
\newblock \showarticletitle{Self-introspective decoding: Alleviating hallucinations for large vision-language models}.
\newblock \bibinfo{journal}{\emph{arXiv preprint arXiv:2408.02032}} (\bibinfo{year}{2024}).
\newblock


\bibitem[Jagannatha and Yu(2020)]%
        {jagannatha-yu-2020-calibrating}
\bibfield{author}{\bibinfo{person}{Abhyuday Jagannatha} {and} \bibinfo{person}{Hong Yu}.} \bibinfo{year}{2020}\natexlab{}.
\newblock \showarticletitle{Calibrating Structured Output Predictors for Natural Language Processing}. In \bibinfo{booktitle}{\emph{Proceedings of the 58th Annual Meeting of the Association for Computational Linguistics}}. \bibinfo{pages}{2078--2092}.
\newblock


\bibitem[Jiang et~al\mbox{.}(2018)]%
        {Jiang2018ToClassifier}
\bibfield{author}{\bibinfo{person}{Heinrich Jiang}, \bibinfo{person}{Been Kim}, \bibinfo{person}{Maya Gupta}, {and} \bibinfo{person}{Melody~Y. Guan}.} \bibinfo{year}{2018}\natexlab{}.
\newblock \showarticletitle{{To trust or not to trust a classifier}}. In \bibinfo{booktitle}{\emph{Advances in Neural Information Processing Systems}}.
\newblock
\showISSN{10495258}


\bibitem[Jiang et~al\mbox{.}(2021)]%
        {jiang-etal-2021-know}
\bibfield{author}{\bibinfo{person}{Zhengbao Jiang}, \bibinfo{person}{Jun Araki}, \bibinfo{person}{Haibo Ding}, {and} \bibinfo{person}{Graham Neubig}.} \bibinfo{year}{2021}\natexlab{}.
\newblock \showarticletitle{How Can We Know When Language Models Know? On the Calibration of Language Models for Question Answering}.
\newblock \bibinfo{journal}{\emph{Transactions of the Association for Computational Linguistics}} (\bibinfo{year}{2021}), \bibinfo{pages}{962--977}.
\newblock


\bibitem[Jin et~al\mbox{.}(2021)]%
        {jin2021disease}
\bibfield{author}{\bibinfo{person}{Di Jin}, \bibinfo{person}{Eileen Pan}, \bibinfo{person}{Nassim Oufattole}, \bibinfo{person}{Wei-Hung Weng}, \bibinfo{person}{Hanyi Fang}, {and} \bibinfo{person}{Peter Szolovits}.} \bibinfo{year}{2021}\natexlab{}.
\newblock \showarticletitle{What disease does this patient have? a large-scale open domain question answering dataset from medical exams}.
\newblock \bibinfo{journal}{\emph{Applied Sciences}} \bibinfo{volume}{11}, \bibinfo{number}{14} (\bibinfo{year}{2021}), \bibinfo{pages}{6421}.
\newblock


\bibitem[Joshi et~al\mbox{.}(2017)]%
        {joshi2017triviaqa}
\bibfield{author}{\bibinfo{person}{Mandar Joshi}, \bibinfo{person}{Eunsol Choi}, \bibinfo{person}{Daniel~S Weld}, {and} \bibinfo{person}{Luke Zettlemoyer}.} \bibinfo{year}{2017}\natexlab{}.
\newblock \showarticletitle{TriviaQA: A Large Scale Distantly Supervised Challenge Dataset for Reading Comprehension}. In \bibinfo{booktitle}{\emph{Proceedings of the 55th Annual Meeting of the Association for Computational Linguistics}}. \bibinfo{pages}{1601--1611}.
\newblock


\bibitem[Kadavath et~al\mbox{.}(2022)]%
        {kadavath2022language}
\bibfield{author}{\bibinfo{person}{Saurav Kadavath}, \bibinfo{person}{Tom Conerly}, \bibinfo{person}{Amanda Askell}, \bibinfo{person}{Tom Henighan}, \bibinfo{person}{Dawn Drain}, \bibinfo{person}{Ethan Perez}, \bibinfo{person}{Nicholas Schiefer}, \bibinfo{person}{Zac Hatfield-Dodds}, \bibinfo{person}{Nova DasSarma}, \bibinfo{person}{Eli Tran-Johnson}, {et~al\mbox{.}}} \bibinfo{year}{2022}\natexlab{}.
\newblock \showarticletitle{Language models (mostly) know what they know}.
\newblock \bibinfo{journal}{\emph{arXiv preprint arXiv:2207.05221}} (\bibinfo{year}{2022}).
\newblock


\bibitem[Kapoor et~al\mbox{.}(2024)]%
        {kapoor-etal-2024-calibration}
\bibfield{author}{\bibinfo{person}{Sanyam Kapoor}, \bibinfo{person}{Nate Gruver}, \bibinfo{person}{Manley Roberts}, \bibinfo{person}{Arka Pal}, \bibinfo{person}{Samuel Dooley}, \bibinfo{person}{Micah Goldblum}, {and} \bibinfo{person}{Andrew Wilson}.} \bibinfo{year}{2024}\natexlab{}.
\newblock \showarticletitle{Calibration-Tuning: Teaching Large Language Models to Know What They Don`t Know}. In \bibinfo{booktitle}{\emph{Proceedings of the 1st Workshop on Uncertainty-Aware NLP (UncertaiNLP 2024)}}. \bibinfo{pages}{1--14}.
\newblock


\bibitem[Kuhn et~al\mbox{.}(2023)]%
        {kuhn2023semantic}
\bibfield{author}{\bibinfo{person}{Lorenz Kuhn}, \bibinfo{person}{Yarin Gal}, {and} \bibinfo{person}{Sebastian Farquhar}.} \bibinfo{year}{2023}\natexlab{}.
\newblock \showarticletitle{Semantic Uncertainty: Linguistic Invariances for Uncertainty Estimation in Natural Language Generation}. In \bibinfo{booktitle}{\emph{The Eleventh International Conference on Learning Representations}}.
\newblock


\bibitem[Kumar and Sarawagi(2019)]%
        {kumar2019calibration}
\bibfield{author}{\bibinfo{person}{Aviral Kumar} {and} \bibinfo{person}{Sunita Sarawagi}.} \bibinfo{year}{2019}\natexlab{}.
\newblock \showarticletitle{Calibration of encoder decoder models for neural machine translation}.
\newblock \bibinfo{journal}{\emph{arXiv preprint arXiv:1903.00802}} (\bibinfo{year}{2019}).
\newblock


\bibitem[Kumar et~al\mbox{.}(2023)]%
        {kumar2023conformal}
\bibfield{author}{\bibinfo{person}{Bhawesh Kumar}, \bibinfo{person}{Charlie Lu}, \bibinfo{person}{Gauri Gupta}, \bibinfo{person}{Anil Palepu}, \bibinfo{person}{David Bellamy}, \bibinfo{person}{Ramesh Raskar}, {and} \bibinfo{person}{Andrew Beam}.} \bibinfo{year}{2023}\natexlab{}.
\newblock \showarticletitle{Conformal prediction with large language models for multi-choice question answering}.
\newblock \bibinfo{journal}{\emph{arXiv preprint arXiv:2305.18404}} (\bibinfo{year}{2023}).
\newblock


\bibitem[Lai et~al\mbox{.}(2017)]%
        {lai2017race}
\bibfield{author}{\bibinfo{person}{Guokun Lai}, \bibinfo{person}{Qizhe Xie}, \bibinfo{person}{Hanxiao Liu}, \bibinfo{person}{Yiming Yang}, {and} \bibinfo{person}{Eduard Hovy}.} \bibinfo{year}{2017}\natexlab{}.
\newblock \showarticletitle{RACE: Large-scale ReAding Comprehension Dataset From Examinations}. In \bibinfo{booktitle}{\emph{Proceedings of the 2017 Conference on Empirical Methods in Natural Language Processing}}. \bibinfo{pages}{785--794}.
\newblock


\bibitem[Lai et~al\mbox{.}(2025)]%
        {lai2023llmlight}
\bibfield{author}{\bibinfo{person}{Siqi Lai}, \bibinfo{person}{Zhao Xu}, \bibinfo{person}{Weijia Zhang}, \bibinfo{person}{Hao Liu}, {and} \bibinfo{person}{Hui Xiong}.} \bibinfo{year}{2025}\natexlab{}.
\newblock \showarticletitle{LLMLight: Large Language Models as Traffic Signal Control Agents}.
\newblock \bibinfo{journal}{\emph{31st ACM SIGKDD Conference on Knowledge Discovery and Data Mining}} (\bibinfo{year}{2025}).
\newblock


\bibitem[Lakshminarayanan et~al\mbox{.}(2017)]%
        {lakshminarayanan2017simple}
\bibfield{author}{\bibinfo{person}{Balaji Lakshminarayanan}, \bibinfo{person}{Alexander Pritzel}, {and} \bibinfo{person}{Charles Blundell}.} \bibinfo{year}{2017}\natexlab{}.
\newblock \showarticletitle{Simple and scalable predictive uncertainty estimation using deep ensembles}.
\newblock \bibinfo{journal}{\emph{Advances in neural information processing systems}}  \bibinfo{volume}{30} (\bibinfo{year}{2017}).
\newblock


\bibitem[Li et~al\mbox{.}(2024a)]%
        {li2024llm}
\bibfield{author}{\bibinfo{person}{Baolin Li}, \bibinfo{person}{Yankai Jiang}, \bibinfo{person}{Vijay Gadepally}, {and} \bibinfo{person}{Devesh Tiwari}.} \bibinfo{year}{2024}\natexlab{a}.
\newblock \showarticletitle{Llm inference serving: Survey of recent advances and opportunities}.
\newblock \bibinfo{journal}{\emph{arXiv preprint arXiv:2407.12391}} (\bibinfo{year}{2024}).
\newblock


\bibitem[Li et~al\mbox{.}(2023a)]%
        {li2023halueval}
\bibfield{author}{\bibinfo{person}{Junyi Li}, \bibinfo{person}{Xiaoxue Cheng}, \bibinfo{person}{Wayne~Xin Zhao}, \bibinfo{person}{Jian-Yun Nie}, {and} \bibinfo{person}{Ji-Rong Wen}.} \bibinfo{year}{2023}\natexlab{a}.
\newblock \showarticletitle{HaluEval: A Large-Scale Hallucination Evaluation Benchmark for Large Language Models}. In \bibinfo{booktitle}{\emph{Proceedings of the 2023 Conference on Empirical Methods in Natural Language Processing}}. \bibinfo{pages}{6449--6464}.
\newblock


\bibitem[Li et~al\mbox{.}(2024b)]%
        {li2024political}
\bibfield{author}{\bibinfo{person}{Lincan Li}, \bibinfo{person}{Jiaqi Li}, \bibinfo{person}{Catherine Chen}, \bibinfo{person}{Fred Gui}, \bibinfo{person}{Hongjia Yang}, \bibinfo{person}{Chenxiao Yu}, \bibinfo{person}{Zhengguang Wang}, \bibinfo{person}{Jianing Cai}, \bibinfo{person}{Junlong~Aaron Zhou}, \bibinfo{person}{Bolin Shen}, {et~al\mbox{.}}} \bibinfo{year}{2024}\natexlab{b}.
\newblock \showarticletitle{Political-llm: Large language models in political science}.
\newblock \bibinfo{journal}{\emph{arXiv preprint arXiv:2412.06864}} (\bibinfo{year}{2024}).
\newblock


\bibitem[Li et~al\mbox{.}(2023b)]%
        {li2023unify}
\bibfield{author}{\bibinfo{person}{Yaowei Li}, \bibinfo{person}{Bang Yang}, \bibinfo{person}{Xuxin Cheng}, \bibinfo{person}{Zhihong Zhu}, \bibinfo{person}{Hongxiang Li}, {and} \bibinfo{person}{Yuexian Zou}.} \bibinfo{year}{2023}\natexlab{b}.
\newblock \showarticletitle{Unify, align and refine: Multi-level semantic alignment for radiology report generation}. In \bibinfo{booktitle}{\emph{Proceedings of the IEEE/CVF international conference on computer vision}}. \bibinfo{pages}{2863--2874}.
\newblock


\bibitem[Li et~al\mbox{.}(2024c)]%
        {li2024uncertaintyrag}
\bibfield{author}{\bibinfo{person}{Zixuan Li}, \bibinfo{person}{Jing Xiong}, \bibinfo{person}{Fanghua Ye}, \bibinfo{person}{Chuanyang Zheng}, \bibinfo{person}{Xun Wu}, \bibinfo{person}{Jianqiao Lu}, \bibinfo{person}{Zhongwei Wan}, \bibinfo{person}{Xiaodan Liang}, \bibinfo{person}{Chengming Li}, \bibinfo{person}{Zhenan Sun}, {et~al\mbox{.}}} \bibinfo{year}{2024}\natexlab{c}.
\newblock \showarticletitle{UncertaintyRAG: Span-Level Uncertainty Enhanced Long-Context Modeling for Retrieval-Augmented Generation}.
\newblock \bibinfo{journal}{\emph{arXiv preprint arXiv:2410.02719}} (\bibinfo{year}{2024}).
\newblock


\bibitem[Liang et~al\mbox{.}(2024b)]%
        {liang2024introspective}
\bibfield{author}{\bibinfo{person}{Kaiqu Liang}, \bibinfo{person}{Zixu Zhang}, {and} \bibinfo{person}{Jaime Fisac}.} \bibinfo{year}{2024}\natexlab{b}.
\newblock \showarticletitle{Introspective Planning: Aligning Robots' Uncertainty with Inherent Task Ambiguity}.
\newblock \bibinfo{journal}{\emph{Advances in Neural Information Processing Systems}}  \bibinfo{volume}{37} (\bibinfo{year}{2024}), \bibinfo{pages}{71998--72031}.
\newblock


\bibitem[Liang et~al\mbox{.}(2024a)]%
        {liang2024revisiting}
\bibfield{author}{\bibinfo{person}{Siyuan Liang}, \bibinfo{person}{Jiawei Liang}, \bibinfo{person}{Tianyu Pang}, \bibinfo{person}{Chao Du}, \bibinfo{person}{Aishan Liu}, \bibinfo{person}{Ee-Chien Chang}, {and} \bibinfo{person}{Xiaochun Cao}.} \bibinfo{year}{2024}\natexlab{a}.
\newblock \showarticletitle{Revisiting backdoor attacks against large vision-language models}.
\newblock \bibinfo{journal}{\emph{arXiv preprint arXiv:2406.18844}} (\bibinfo{year}{2024}).
\newblock


\bibitem[Lin et~al\mbox{.}(2022a)]%
        {lin2022teaching}
\bibfield{author}{\bibinfo{person}{Stephanie Lin}, \bibinfo{person}{Jacob Hilton}, {and} \bibinfo{person}{Owain Evans}.} \bibinfo{year}{2022}\natexlab{a}.
\newblock \showarticletitle{Teaching models to express their uncertainty in words}.
\newblock \bibinfo{journal}{\emph{arXiv preprint arXiv:2205.14334}} (\bibinfo{year}{2022}).
\newblock


\bibitem[Lin et~al\mbox{.}(2022b)]%
        {lin2021truthfulqa}
\bibfield{author}{\bibinfo{person}{Stephanie Lin}, \bibinfo{person}{Jacob Hilton}, {and} \bibinfo{person}{Owain Evans}.} \bibinfo{year}{2022}\natexlab{b}.
\newblock \showarticletitle{TruthfulQA: Measuring How Models Mimic Human Falsehoods}. In \bibinfo{booktitle}{\emph{Proceedings of the 60th Annual Meeting of the Association for Computational Linguistics (Volume 1: Long Papers)}}. \bibinfo{pages}{3214--3252}.
\newblock


\bibitem[Lin et~al\mbox{.}(2023)]%
        {lin2023generating}
\bibfield{author}{\bibinfo{person}{Zhen Lin}, \bibinfo{person}{Shubhendu Trivedi}, {and} \bibinfo{person}{Jimeng Sun}.} \bibinfo{year}{2023}\natexlab{}.
\newblock \showarticletitle{Generating with Confidence: Uncertainty Quantification for Black-box Large Language Models}.
\newblock \bibinfo{journal}{\emph{Transactions on Machine Learning Research}} (\bibinfo{year}{2023}).
\newblock


\bibitem[Lin et~al\mbox{.}(2024)]%
        {lin2024contextualized}
\bibfield{author}{\bibinfo{person}{Zhen Lin}, \bibinfo{person}{Shubhendu Trivedi}, {and} \bibinfo{person}{Jimeng Sun}.} \bibinfo{year}{2024}\natexlab{}.
\newblock \showarticletitle{Contextualized Sequence Likelihood: Enhanced Confidence Scores for Natural Language Generation}. In \bibinfo{booktitle}{\emph{Proceedings of the 2024 Conference on Empirical Methods in Natural Language Processing}}. \bibinfo{pages}{10351--10368}.
\newblock


\bibitem[Ling et~al\mbox{.}(2024)]%
        {ling2024uncertainty}
\bibfield{author}{\bibinfo{person}{Chen Ling}, \bibinfo{person}{Xujiang Zhao}, \bibinfo{person}{Xuchao Zhang}, \bibinfo{person}{Wei Cheng}, \bibinfo{person}{Yanchi Liu}, \bibinfo{person}{Yiyou Sun}, \bibinfo{person}{Mika Oishi}, \bibinfo{person}{Takao Osaki}, \bibinfo{person}{Katsushi Matsuda}, \bibinfo{person}{Jie Ji}, \bibinfo{person}{Guangji Bai}, \bibinfo{person}{Liang Zhao}, {and} \bibinfo{person}{Haifeng Chen}.} \bibinfo{year}{2024}\natexlab{}.
\newblock \showarticletitle{Uncertainty Quantification for In-Context Learning of Large Language Models}. In \bibinfo{booktitle}{\emph{Proceedings of the 2024 Conference of the North American Chapter of the Association for Computational Linguistics: Human Language Technologies (Volume 1: Long Papers)}}. \bibinfo{pages}{3357--3370}.
\newblock


\bibitem[Liu et~al\mbox{.}(2024c)]%
        {liu2024uncertainty}
\bibfield{author}{\bibinfo{person}{Linyu Liu}, \bibinfo{person}{Yu Pan}, \bibinfo{person}{Xiaocheng Li}, {and} \bibinfo{person}{Guanting Chen}.} \bibinfo{year}{2024}\natexlab{c}.
\newblock \showarticletitle{Uncertainty estimation and quantification for llms: A simple supervised approach}.
\newblock \bibinfo{journal}{\emph{arXiv preprint arXiv:2404.15993}} (\bibinfo{year}{2024}).
\newblock


\bibitem[Liu et~al\mbox{.}(2024b)]%
        {liu2024can}
\bibfield{author}{\bibinfo{person}{Shudong Liu}, \bibinfo{person}{Zhaocong Li}, \bibinfo{person}{Xuebo Liu}, \bibinfo{person}{Runzhe Zhan}, \bibinfo{person}{Derek Wong}, \bibinfo{person}{Lidia Chao}, {and} \bibinfo{person}{Min Zhang}.} \bibinfo{year}{2024}\natexlab{b}.
\newblock \showarticletitle{Can LLMs learn uncertainty on their own? expressing uncertainty effectively in a self-training manner}. In \bibinfo{booktitle}{\emph{Proceedings of the 2024 Conference on Empirical Methods in Natural Language Processing}}. \bibinfo{pages}{21635--21645}.
\newblock


\bibitem[Liu et~al\mbox{.}(2024a)]%
        {liu2024litcab}
\bibfield{author}{\bibinfo{person}{Xin Liu}, \bibinfo{person}{Muhammad Khalifa}, {and} \bibinfo{person}{Lu Wang}.} \bibinfo{year}{2024}\natexlab{a}.
\newblock \showarticletitle{LitCab: Lightweight Language Model Calibration over Short- and Long-form Responses}. In \bibinfo{booktitle}{\emph{The Twelfth International Conference on Learning Representations}}.
\newblock


\bibitem[Liu et~al\mbox{.}(2025)]%
        {liu2025mcqa}
\bibfield{author}{\bibinfo{person}{Xiaoou Liu}, \bibinfo{person}{Zhen Lin}, \bibinfo{person}{Longchao Da}, \bibinfo{person}{Chacha Chen}, \bibinfo{person}{Shubhendu Trivedi}, {and} \bibinfo{person}{Hua Wei}.} \bibinfo{year}{2025}\natexlab{}.
\newblock \showarticletitle{MCQA-Eval: Efficient Confidence Evaluation in NLG with Gold-Standard Correctness Labels}.
\newblock \bibinfo{journal}{\emph{arXiv preprint arXiv:2502.14268}} (\bibinfo{year}{2025}).
\newblock


\bibitem[Lu et~al\mbox{.}(2024)]%
        {lu2024merge}
\bibfield{author}{\bibinfo{person}{Jinliang Lu}, \bibinfo{person}{Ziliang Pang}, \bibinfo{person}{Min Xiao}, \bibinfo{person}{Yaochen Zhu}, \bibinfo{person}{Rui Xia}, {and} \bibinfo{person}{Jiajun Zhang}.} \bibinfo{year}{2024}\natexlab{}.
\newblock \showarticletitle{Merge, ensemble, and cooperate! a survey on collaborative strategies in the era of large language models}.
\newblock \bibinfo{journal}{\emph{arXiv preprint arXiv:2407.06089}} (\bibinfo{year}{2024}).
\newblock


\bibitem[Madaan et~al\mbox{.}(2023)]%
        {madaan2023self}
\bibfield{author}{\bibinfo{person}{Aman Madaan}, \bibinfo{person}{Niket Tandon}, \bibinfo{person}{Prakhar Gupta}, \bibinfo{person}{Skyler Hallinan}, \bibinfo{person}{Luyu Gao}, \bibinfo{person}{Sarah Wiegreffe}, \bibinfo{person}{Uri Alon}, \bibinfo{person}{Nouha Dziri}, \bibinfo{person}{Shrimai Prabhumoye}, \bibinfo{person}{Yiming Yang}, {et~al\mbox{.}}} \bibinfo{year}{2023}\natexlab{}.
\newblock \showarticletitle{Self-refine: Iterative refinement with self-feedback}.
\newblock \bibinfo{journal}{\emph{Advances in Neural Information Processing Systems}}  \bibinfo{volume}{36} (\bibinfo{year}{2023}), \bibinfo{pages}{46534--46594}.
\newblock


\bibitem[Malinin and Gales(2021)]%
        {malinin2021uncertainty}
\bibfield{author}{\bibinfo{person}{Andrey Malinin} {and} \bibinfo{person}{Mark Gales}.} \bibinfo{year}{2021}\natexlab{}.
\newblock \showarticletitle{Uncertainty Estimation in Autoregressive Structured Prediction}. In \bibinfo{booktitle}{\emph{International Conference on Learning Representations}}.
\newblock


\bibitem[Manakul et~al\mbox{.}({[n.\,d.]})]%
        {manakul2023selfcheckgpt}
\bibfield{author}{\bibinfo{person}{Potsawee Manakul}, \bibinfo{person}{Adian Liusie}, {and} \bibinfo{person}{Mark Gales}.} \bibinfo{year}{[n.\,d.]}\natexlab{}.
\newblock \showarticletitle{SelfCheckGPT: Zero-Resource Black-Box Hallucination Detection for Generative Large Language Models}. In \bibinfo{booktitle}{\emph{2023 Conference on Empirical Methods in Natural Language Processing}}.
\newblock


\bibitem[Margatina et~al\mbox{.}(2023)]%
        {margatina2023active}
\bibfield{author}{\bibinfo{person}{Katerina Margatina}, \bibinfo{person}{Timo Schick}, \bibinfo{person}{Nikolaos Aletras}, {and} \bibinfo{person}{Jane Dwivedi-Yu}.} \bibinfo{year}{2023}\natexlab{}.
\newblock \showarticletitle{Active Learning Principles for In-Context Learning with Large Language Models}. In \bibinfo{booktitle}{\emph{Findings of the Association for Computational Linguistics}}. \bibinfo{pages}{5011--5034}.
\newblock


\bibitem[Min et~al\mbox{.}(2023)]%
        {min2023factscore}
\bibfield{author}{\bibinfo{person}{Sewon Min}, \bibinfo{person}{Kalpesh Krishna}, \bibinfo{person}{Xinxi Lyu}, \bibinfo{person}{Mike Lewis}, \bibinfo{person}{Wen-tau Yih}, \bibinfo{person}{Pang Koh}, \bibinfo{person}{Mohit yyer}, \bibinfo{person}{Luke Zettlemoyer}, {and} \bibinfo{person}{Hannaneh Hajishirzi}.} \bibinfo{year}{2023}\natexlab{}.
\newblock \showarticletitle{{FA}ct{S}core: Fine-grained Atomic Evaluation of Factual Precision in Long Form Text Generation}. In \bibinfo{booktitle}{\emph{Proceedings of the 2023 Conference on Empirical Methods in Natural Language Processing}}. \bibinfo{pages}{12076--12100}.
\newblock


\bibitem[Min et~al\mbox{.}(2020)]%
        {min2020ambigqa}
\bibfield{author}{\bibinfo{person}{Sewon Min}, \bibinfo{person}{Julian Michael}, \bibinfo{person}{Hannaneh Hajishirzi}, {and} \bibinfo{person}{Luke Zettlemoyer}.} \bibinfo{year}{2020}\natexlab{}.
\newblock \showarticletitle{{A}mbig{QA}: Answering Ambiguous Open-domain Questions}. In \bibinfo{booktitle}{\emph{EMNLP}}.
\newblock


\bibitem[Mo and Xin(2024)]%
        {mo2024tree}
\bibfield{author}{\bibinfo{person}{Shentong Mo} {and} \bibinfo{person}{Miao Xin}.} \bibinfo{year}{2024}\natexlab{}.
\newblock \showarticletitle{Tree of uncertain thoughts reasoning for large language models}. In \bibinfo{booktitle}{\emph{ICASSP 2024-2024 IEEE International Conference on Acoustics, Speech and Signal Processing (ICASSP)}}. IEEE, \bibinfo{pages}{12742--12746}.
\newblock


\bibitem[Mondorf and Plank(2024)]%
        {mondorfbeyond}
\bibfield{author}{\bibinfo{person}{Philipp Mondorf} {and} \bibinfo{person}{Barbara Plank}.} \bibinfo{year}{2024}\natexlab{}.
\newblock \showarticletitle{Beyond Accuracy: Evaluating the Reasoning Behavior of Large Language Models-A Survey}. In \bibinfo{booktitle}{\emph{First Conference on Language Modeling}}.
\newblock


\bibitem[Mora-Cross and Calderon-Ramirez(2024)]%
        {mora2024uncertainty}
\bibfield{author}{\bibinfo{person}{Maria Mora-Cross} {and} \bibinfo{person}{Saul Calderon-Ramirez}.} \bibinfo{year}{2024}\natexlab{}.
\newblock \showarticletitle{Uncertainty estimation in large language models to support biodiversity conservation}. In \bibinfo{booktitle}{\emph{Proceedings of the 2024 Conference of the North American Chapter of the Association for Computational Linguistics: Human Language Technologies (Volume 6: Industry Track)}}. \bibinfo{pages}{368--378}.
\newblock


\bibitem[Mukherjee and Awadallah(2020)]%
        {mukherjee2020uncertainty}
\bibfield{author}{\bibinfo{person}{Subhabrata Mukherjee} {and} \bibinfo{person}{Ahmed Awadallah}.} \bibinfo{year}{2020}\natexlab{}.
\newblock \showarticletitle{Uncertainty-aware self-training for few-shot text classification}.
\newblock \bibinfo{journal}{\emph{Advances in Neural Information Processing Systems}}  \bibinfo{volume}{33} (\bibinfo{year}{2020}), \bibinfo{pages}{21199--21212}.
\newblock


\bibitem[Mullen~Jr and Manocha(2024)]%
        {mullen2024lap}
\bibfield{author}{\bibinfo{person}{James~F Mullen~Jr} {and} \bibinfo{person}{Dinesh Manocha}.} \bibinfo{year}{2024}\natexlab{}.
\newblock \showarticletitle{LAP, Using Action Feasibility for Improved Uncertainty Alignment of Large Language Model Planners}.
\newblock \bibinfo{journal}{\emph{arXiv preprint arXiv:2403.13198}} (\bibinfo{year}{2024}).
\newblock


\bibitem[Murphy and Winkler(1977)]%
        {Murphy1977}
\bibfield{author}{\bibinfo{person}{Allan~H. Murphy} {and} \bibinfo{person}{Robert~L. Winkler}.} \bibinfo{year}{1977}\natexlab{}.
\newblock \showarticletitle{Reliability of Subjective Probability Forecasts of Precipitation and Temperature}.
\newblock \bibinfo{journal}{\emph{Journal of The Royal Statistical Society Series C-applied Statistics}}  \bibinfo{volume}{26} (\bibinfo{year}{1977}), \bibinfo{pages}{41--47}.
\newblock


\bibitem[Nadeem et~al\mbox{.}(2009)]%
        {nadeem2009accuracy}
\bibfield{author}{\bibinfo{person}{Malik Sajjad~Ahmed Nadeem}, \bibinfo{person}{Jean-Daniel Zucker}, {and} \bibinfo{person}{Blaise Hanczar}.} \bibinfo{year}{2009}\natexlab{}.
\newblock \showarticletitle{Accuracy-rejection curves (ARCs) for comparing classification methods with a reject option}. In \bibinfo{booktitle}{\emph{Machine Learning in Systems Biology}}. \bibinfo{pages}{65--81}.
\newblock


\bibitem[Nikitin et~al\mbox{.}(2024)]%
        {nikitin2024kernel}
\bibfield{author}{\bibinfo{person}{Alexander Nikitin}, \bibinfo{person}{Jannik Kossen}, \bibinfo{person}{Yarin Gal}, {and} \bibinfo{person}{Pekka Marttinen}.} \bibinfo{year}{2024}\natexlab{}.
\newblock \showarticletitle{Kernel language entropy: Fine-grained uncertainty quantification for LLMs from semantic similarities}.
\newblock \bibinfo{journal}{\emph{Advances in Neural Information Processing Systems}}  \bibinfo{volume}{37} (\bibinfo{year}{2024}), \bibinfo{pages}{8901--8929}.
\newblock


\bibitem[Ong et~al\mbox{.}(2024)]%
        {ong2024simple}
\bibfield{author}{\bibinfo{person}{Hyobin Ong}, \bibinfo{person}{Youngwoo Yoon}, \bibinfo{person}{Jaewoo Choi}, {and} \bibinfo{person}{Minsu Jang}.} \bibinfo{year}{2024}\natexlab{}.
\newblock \showarticletitle{A Simple Baseline for Uncertainty-Aware Language-Oriented Task Planner for Embodied Agents}. In \bibinfo{booktitle}{\emph{2024 21st International Conference on Ubiquitous Robots (UR)}}. \bibinfo{pages}{677--682}.
\newblock


\bibitem[Panickssery et~al\mbox{.}(2024)]%
        {panickssery2024llm}
\bibfield{author}{\bibinfo{person}{Arjun Panickssery}, \bibinfo{person}{Samuel Bowman}, {and} \bibinfo{person}{Shi Feng}.} \bibinfo{year}{2024}\natexlab{}.
\newblock \showarticletitle{Llm evaluators recognize and favor their own generations}.
\newblock \bibinfo{journal}{\emph{Advances in Neural Information Processing Systems}}  \bibinfo{volume}{37} (\bibinfo{year}{2024}), \bibinfo{pages}{68772--68802}.
\newblock


\bibitem[Papineni et~al\mbox{.}(2002)]%
        {papineni2002bleu}
\bibfield{author}{\bibinfo{person}{Kishore Papineni}, \bibinfo{person}{Salim Roukos}, \bibinfo{person}{Todd Ward}, {and} \bibinfo{person}{Wei-Jing Zhu}.} \bibinfo{year}{2002}\natexlab{}.
\newblock \showarticletitle{Bleu: a method for automatic evaluation of machine translation}. In \bibinfo{booktitle}{\emph{Proceedings of the 40th annual meeting of the Association for Computational Linguistics}}. \bibinfo{pages}{311--318}.
\newblock


\bibitem[Qiu et~al\mbox{.}(2024)]%
        {qiu2024llm}
\bibfield{author}{\bibinfo{person}{Jianing Qiu}, \bibinfo{person}{Kyle Lam}, \bibinfo{person}{Guohao Li}, \bibinfo{person}{Amish Acharya}, \bibinfo{person}{Tien~Yin Wong}, \bibinfo{person}{Ara Darzi}, \bibinfo{person}{Wu Yuan}, {and} \bibinfo{person}{Eric~J Topol}.} \bibinfo{year}{2024}\natexlab{}.
\newblock \showarticletitle{LLM-based agentic systems in medicine and healthcare}.
\newblock \bibinfo{journal}{\emph{Nature Machine Intelligence}} \bibinfo{volume}{6}, \bibinfo{number}{12} (\bibinfo{year}{2024}), \bibinfo{pages}{1418--1420}.
\newblock


\bibitem[Quach et~al\mbox{.}({[n.\,d.]})]%
        {quachconformal}
\bibfield{author}{\bibinfo{person}{Victor Quach}, \bibinfo{person}{Adam Fisch}, \bibinfo{person}{Tal Schuster}, \bibinfo{person}{Adam Yala}, \bibinfo{person}{Jae~Ho Sohn}, \bibinfo{person}{Tommi~S Jaakkola}, {and} \bibinfo{person}{Regina Barzilay}.} \bibinfo{year}{[n.\,d.]}\natexlab{}.
\newblock \showarticletitle{Conformal Language Modeling}. In \bibinfo{booktitle}{\emph{The Twelfth International Conference on Learning Representations}}.
\newblock


\bibitem[Rajpurkar et~al\mbox{.}(2016)]%
        {rajpurkar2016squad}
\bibfield{author}{\bibinfo{person}{Pranav Rajpurkar}, \bibinfo{person}{Jian Zhang}, \bibinfo{person}{Konstantin Lopyrev}, {and} \bibinfo{person}{Percy Liang}.} \bibinfo{year}{2016}\natexlab{}.
\newblock \showarticletitle{Squad: 100,000+ questions for machine comprehension of text}.
\newblock \bibinfo{journal}{\emph{arXiv preprint arXiv:1606.05250}} (\bibinfo{year}{2016}).
\newblock


\bibitem[Reddy et~al\mbox{.}(2019)]%
        {reddy2019coqa}
\bibfield{author}{\bibinfo{person}{Siva Reddy}, \bibinfo{person}{Danqi Chen}, {and} \bibinfo{person}{Christopher~D Manning}.} \bibinfo{year}{2019}\natexlab{}.
\newblock \showarticletitle{Coqa: A conversational question answering challenge}.
\newblock \bibinfo{journal}{\emph{Transactions of the Association for Computational Linguistics}}  \bibinfo{volume}{7} (\bibinfo{year}{2019}), \bibinfo{pages}{249--266}.
\newblock


\bibitem[Rein et~al\mbox{.}(2024)]%
        {rein2024gpqa}
\bibfield{author}{\bibinfo{person}{David Rein}, \bibinfo{person}{Betty~Li Hou}, \bibinfo{person}{Asa~Cooper Stickland}, \bibinfo{person}{Jackson Petty}, \bibinfo{person}{Richard~Yuanzhe Pang}, \bibinfo{person}{Julien Dirani}, \bibinfo{person}{Julian Michael}, {and} \bibinfo{person}{Samuel~R Bowman}.} \bibinfo{year}{2024}\natexlab{}.
\newblock \showarticletitle{Gpqa: A graduate-level google-proof q\&a benchmark}. In \bibinfo{booktitle}{\emph{First Conference on Language Modeling}}.
\newblock


\bibitem[Ren et~al\mbox{.}(2023a)]%
        {ren2022out}
\bibfield{author}{\bibinfo{person}{Jie Ren}, \bibinfo{person}{Jiaming Luo}, \bibinfo{person}{Yao Zhao}, \bibinfo{person}{Kundan Krishna}, \bibinfo{person}{Mohammad Saleh}, \bibinfo{person}{Balaji Lakshminarayanan}, {and} \bibinfo{person}{Peter~J Liu}.} \bibinfo{year}{2023}\natexlab{a}.
\newblock \showarticletitle{Out-of-Distribution Detection and Selective Generation for Conditional Language Models}. In \bibinfo{booktitle}{\emph{The Eleventh International Conference on Learning Representations}}.
\newblock


\bibitem[Ren et~al\mbox{.}(2023b)]%
        {pmlr-v239-ren23a}
\bibfield{author}{\bibinfo{person}{Jie Ren}, \bibinfo{person}{Yao Zhao}, \bibinfo{person}{Tu Vu}, \bibinfo{person}{Peter~J. Liu}, {and} \bibinfo{person}{Balaji Lakshminarayanan}.} \bibinfo{year}{2023}\natexlab{b}.
\newblock \showarticletitle{Self-Evaluation Improves Selective Generation in Large Language Models}. In \bibinfo{booktitle}{\emph{Proceedings on "I Can't Believe It's Not Better: Failure Modes in the Age of Foundation Models" at NeurIPS 2023 Workshops}}, Vol.~\bibinfo{volume}{239}. \bibinfo{pages}{49--64}.
\newblock


\bibitem[Savage et~al\mbox{.}(2024)]%
        {savage2024large}
\bibfield{author}{\bibinfo{person}{Thomas Savage}, \bibinfo{person}{John Wang}, \bibinfo{person}{Robert Gallo}, \bibinfo{person}{Abdessalem Boukil}, \bibinfo{person}{Vishwesh Patel}, \bibinfo{person}{Seyed~Amir Ahmad Safavi-Naini}, \bibinfo{person}{Ali Soroush}, {and} \bibinfo{person}{Jonathan~H Chen}.} \bibinfo{year}{2024}\natexlab{}.
\newblock \showarticletitle{Large language model uncertainty measurement and calibration for medical diagnosis and treatment}.
\newblock \bibinfo{journal}{\emph{medRxiv}} (\bibinfo{year}{2024}), \bibinfo{pages}{2024--06}.
\newblock


\bibitem[Sen et~al\mbox{.}(2024)]%
        {sen2024erg}
\bibfield{author}{\bibinfo{person}{Sagar Sen}, \bibinfo{person}{Victor Gonzalez}, \bibinfo{person}{Erik~Johannes Husom}, \bibinfo{person}{Simeon Tverdal}, \bibinfo{person}{Shukun Tokas}, {and} \bibinfo{person}{Svein~O Tj{\o}svoll}.} \bibinfo{year}{2024}\natexlab{}.
\newblock \showarticletitle{ERG-AI: enhancing occupational ergonomics with uncertainty-aware ML and LLM feedback}.
\newblock \bibinfo{journal}{\emph{Applied Intelligence}} \bibinfo{volume}{54}, \bibinfo{number}{23} (\bibinfo{year}{2024}), \bibinfo{pages}{12128--12155}.
\newblock


\bibitem[Sensoy et~al\mbox{.}(2018)]%
        {sensoy2018evidential}
\bibfield{author}{\bibinfo{person}{Murat Sensoy}, \bibinfo{person}{Lance Kaplan}, {and} \bibinfo{person}{Melih Kandemir}.} \bibinfo{year}{2018}\natexlab{}.
\newblock \showarticletitle{Evidential deep learning to quantify classification uncertainty}.
\newblock \bibinfo{journal}{\emph{Advances in neural information processing systems}}  \bibinfo{volume}{31} (\bibinfo{year}{2018}).
\newblock


\bibitem[Shafer and Vovk(2008)]%
        {shafer2008tutorial}
\bibfield{author}{\bibinfo{person}{Glenn Shafer} {and} \bibinfo{person}{Vladimir Vovk}.} \bibinfo{year}{2008}\natexlab{}.
\newblock \showarticletitle{A tutorial on conformal prediction.}
\newblock \bibinfo{journal}{\emph{Journal of Machine Learning Research}} \bibinfo{volume}{9}, \bibinfo{number}{3} (\bibinfo{year}{2008}).
\newblock


\bibitem[Shorinwa et~al\mbox{.}(2024)]%
        {shorinwa2024survey}
\bibfield{author}{\bibinfo{person}{Ola Shorinwa}, \bibinfo{person}{Zhiting Mei}, \bibinfo{person}{Justin Lidard}, \bibinfo{person}{Allen~Z Ren}, {and} \bibinfo{person}{Anirudha Majumdar}.} \bibinfo{year}{2024}\natexlab{}.
\newblock \showarticletitle{A survey on uncertainty quantification of large language models: Taxonomy, open research challenges, and future directions}.
\newblock \bibinfo{journal}{\emph{arXiv preprint arXiv:2412.05563}} (\bibinfo{year}{2024}).
\newblock


\bibitem[Stengel-Eskin and Van~Durme(2023)]%
        {stengel-eskin-van-durme-2023-calibrated}
\bibfield{author}{\bibinfo{person}{Elias Stengel-Eskin} {and} \bibinfo{person}{Benjamin Van~Durme}.} \bibinfo{year}{2023}\natexlab{}.
\newblock \showarticletitle{Calibrated Interpretation: Confidence Estimation in Semantic Parsing}.
\newblock \bibinfo{journal}{\emph{Transactions of the Association for Computational Linguistics}}  \bibinfo{volume}{11} (\bibinfo{year}{2023}), \bibinfo{pages}{1213--1231}.
\newblock


\bibitem[Su et~al\mbox{.}(2024)]%
        {su2024api}
\bibfield{author}{\bibinfo{person}{Jiayuan Su}, \bibinfo{person}{Jing Luo}, \bibinfo{person}{Hongwei Wang}, {and} \bibinfo{person}{Lu Cheng}.} \bibinfo{year}{2024}\natexlab{}.
\newblock \showarticletitle{API Is Enough: Conformal Prediction for Large Language Models Without Logit-Access}. In \bibinfo{booktitle}{\emph{Findings of the Association for Computational Linguistics: EMNLP 2024}}. \bibinfo{pages}{979--995}.
\newblock


\bibitem[Suo et~al\mbox{.}(2025)]%
        {suo2025octopus}
\bibfield{author}{\bibinfo{person}{Wei Suo}, \bibinfo{person}{Lijun Zhang}, \bibinfo{person}{Mengyang Sun}, \bibinfo{person}{Lin~Yuanbo Wu}, \bibinfo{person}{Peng Wang}, {and} \bibinfo{person}{Yanning Zhang}.} \bibinfo{year}{2025}\natexlab{}.
\newblock \showarticletitle{Octopus: Alleviating Hallucination via Dynamic Contrastive Decoding}.
\newblock \bibinfo{journal}{\emph{arXiv preprint arXiv:2503.00361}} (\bibinfo{year}{2025}).
\newblock


\bibitem[Thorne et~al\mbox{.}(2018)]%
        {thorne2018fever}
\bibfield{author}{\bibinfo{person}{James Thorne}, \bibinfo{person}{Andreas Vlachos}, \bibinfo{person}{Christos Christodoulopoulos}, {and} \bibinfo{person}{Arpit Mittal}.} \bibinfo{year}{2018}\natexlab{}.
\newblock \showarticletitle{FEVER: a large-scale dataset for fact extraction and VERification}.
\newblock \bibinfo{journal}{\emph{arXiv preprint arXiv:1803.05355}} (\bibinfo{year}{2018}).
\newblock


\bibitem[Tian et~al\mbox{.}(2023)]%
        {tian-etal-2023-just}
\bibfield{author}{\bibinfo{person}{Katherine Tian}, \bibinfo{person}{Eric Mitchell}, \bibinfo{person}{Allan Zhou}, \bibinfo{person}{Archit Sharma}, \bibinfo{person}{Rafael Rafailov}, \bibinfo{person}{Huaxiu Yao}, \bibinfo{person}{Chelsea Finn}, {and} \bibinfo{person}{Christopher Manning}.} \bibinfo{year}{2023}\natexlab{}.
\newblock \showarticletitle{Just Ask for Calibration: Strategies for Eliciting Calibrated Confidence Scores from Language Models Fine-Tuned with Human Feedback}. In \bibinfo{booktitle}{\emph{Proceedings of the 2023 Conference on Empirical Methods in Natural Language Processing}}. \bibinfo{pages}{5433--5442}.
\newblock


\bibitem[Tsai et~al\mbox{.}(2024)]%
        {tsai2024efficient}
\bibfield{author}{\bibinfo{person}{Yao-Hung~Hubert Tsai}, \bibinfo{person}{Walter Talbott}, {and} \bibinfo{person}{Jian Zhang}.} \bibinfo{year}{2024}\natexlab{}.
\newblock \showarticletitle{Efficient Non-Parametric Uncertainty Quantification for Black-Box Large Language Models and Decision Planning}.
\newblock \bibinfo{journal}{\emph{arXiv preprint arXiv:2402.00251}} (\bibinfo{year}{2024}).
\newblock


\bibitem[Ulmer et~al\mbox{.}(2024)]%
        {ulmer-etal-2024-calibrating}
\bibfield{author}{\bibinfo{person}{Dennis Ulmer}, \bibinfo{person}{Martin Gubri}, \bibinfo{person}{Hwaran Lee}, \bibinfo{person}{Sangdoo Yun}, {and} \bibinfo{person}{Seong Oh}.} \bibinfo{year}{2024}\natexlab{}.
\newblock \showarticletitle{Calibrating Large Language Models Using Their Generations Only}. In \bibinfo{booktitle}{\emph{Proceedings of the 62nd Annual Meeting of the Association for Computational Linguistics (Volume 1: Long Papers)}}. \bibinfo{pages}{15440--15459}.
\newblock


\bibitem[Vashurin et~al\mbox{.}(2025)]%
        {vashurin2024benchmarking}
\bibfield{author}{\bibinfo{person}{Roman Vashurin}, \bibinfo{person}{Ekaterina Fadeeva}, \bibinfo{person}{Artem Vazhentsev}, \bibinfo{person}{Lyudmila Rvanova}, \bibinfo{person}{Daniil Vasilev}, \bibinfo{person}{Akim Tsvigun}, \bibinfo{person}{Sergey Petrakov}, \bibinfo{person}{Rui Xing}, \bibinfo{person}{Abdelrahman Sadallah}, \bibinfo{person}{Kirill Grishchenkov}, \bibinfo{person}{Alexander Panchenko}, \bibinfo{person}{Timothy Baldwin}, \bibinfo{person}{Preslav Nakov}, \bibinfo{person}{Maxim Panov}, {and} \bibinfo{person}{Artem Shelmanov}.} \bibinfo{year}{2025}\natexlab{}.
\newblock \showarticletitle{Benchmarking Uncertainty Quantification Methods for Large Language Models with LM-Polygraph}.
\newblock \bibinfo{journal}{\emph{Transactions of the Association for Computational Linguistics}} (\bibinfo{year}{2025}), \bibinfo{pages}{220--248}.
\newblock


\bibitem[Vaswani et~al\mbox{.}(2017)]%
        {vaswani2017attention}
\bibfield{author}{\bibinfo{person}{Ashish Vaswani}, \bibinfo{person}{Noam Shazeer}, \bibinfo{person}{Niki Parmar}, \bibinfo{person}{Jakob Uszkoreit}, \bibinfo{person}{Llion Jones}, \bibinfo{person}{Aidan~N Gomez}, \bibinfo{person}{{\L}ukasz Kaiser}, {and} \bibinfo{person}{Illia Polosukhin}.} \bibinfo{year}{2017}\natexlab{}.
\newblock \showarticletitle{Attention is all you need}.
\newblock \bibinfo{journal}{\emph{Advances in neural information processing systems}}  \bibinfo{volume}{30} (\bibinfo{year}{2017}).
\newblock


\bibitem[Vazhentsev et~al\mbox{.}(2023)]%
        {vazhentsev2023huq}
\bibfield{author}{\bibinfo{person}{Artem Vazhentsev}, \bibinfo{person}{Gleb Kuzmin}, \bibinfo{person}{Akim Tsvigun}, \bibinfo{person}{Alexander Panchenko}, \bibinfo{person}{Maxim Panov}, \bibinfo{person}{Mikhail Burtsev}, {and} \bibinfo{person}{Artem Shelmanov}.} \bibinfo{year}{2023}\natexlab{}.
\newblock \showarticletitle{Hybrid uncertainty quantification for selective text classification in ambiguous tasks}. In \bibinfo{booktitle}{\emph{Proceedings of the 61st Annual Meeting of the Association for Computational Linguistics (Volume 1: Long Papers)}}. \bibinfo{pages}{11659--11681}.
\newblock


\bibitem[Wang et~al\mbox{.}(2024b)]%
        {wang2024grokking}
\bibfield{author}{\bibinfo{person}{Boshi Wang}, \bibinfo{person}{Xiang Yue}, \bibinfo{person}{Yu Su}, {and} \bibinfo{person}{Huan Sun}.} \bibinfo{year}{2024}\natexlab{b}.
\newblock \showarticletitle{Grokking of implicit reasoning in transformers: A mechanistic journey to the edge of generalization}.
\newblock \bibinfo{journal}{\emph{Advances in Neural Information Processing Systems}}  \bibinfo{volume}{37} (\bibinfo{year}{2024}), \bibinfo{pages}{95238--95265}.
\newblock


\bibitem[Wang et~al\mbox{.}(2020)]%
        {wang-etal-2020-inference}
\bibfield{author}{\bibinfo{person}{Shuo Wang}, \bibinfo{person}{Zhaopeng Tu}, \bibinfo{person}{Shuming Shi}, {and} \bibinfo{person}{Yang Liu}.} \bibinfo{year}{2020}\natexlab{}.
\newblock \showarticletitle{On the Inference Calibration of Neural Machine Translation}. In \bibinfo{booktitle}{\emph{Proceedings of the 58th Annual Meeting of the Association for Computational Linguistics}}.
\newblock


\bibitem[Wang et~al\mbox{.}(2025)]%
        {wang2025blob}
\bibfield{author}{\bibinfo{person}{Yibin Wang}, \bibinfo{person}{Haizhou Shi}, \bibinfo{person}{Ligong Han}, \bibinfo{person}{Dimitris Metaxas}, {and} \bibinfo{person}{Hao Wang}.} \bibinfo{year}{2025}\natexlab{}.
\newblock \showarticletitle{Blob: Bayesian low-rank adaptation by backpropagation for large language models}.
\newblock \bibinfo{journal}{\emph{Advances in Neural Information Processing Systems}}  \bibinfo{volume}{37} (\bibinfo{year}{2025}), \bibinfo{pages}{67758--67794}.
\newblock


\bibitem[Wang et~al\mbox{.}(2024a)]%
        {wang2024conu}
\bibfield{author}{\bibinfo{person}{Zhiyuan Wang}, \bibinfo{person}{Jinhao Duan}, \bibinfo{person}{Lu Cheng}, \bibinfo{person}{Yue Zhang}, \bibinfo{person}{Qingni Wang}, \bibinfo{person}{Xiaoshuang Shi}, \bibinfo{person}{Kaidi Xu}, \bibinfo{person}{Heng~Tao Shen}, {and} \bibinfo{person}{Xiaofeng Zhu}.} \bibinfo{year}{2024}\natexlab{a}.
\newblock \showarticletitle{ConU: Conformal Uncertainty in Large Language Models with Correctness Coverage Guarantees}. In \bibinfo{booktitle}{\emph{Findings of the Association for Computational Linguistics}}. \bibinfo{pages}{6886--6898}.
\newblock


\bibitem[Wu et~al\mbox{.}(2024)]%
        {wu2024uncertainty}
\bibfield{author}{\bibinfo{person}{Jiaxin Wu}, \bibinfo{person}{Yizhou Yu}, {and} \bibinfo{person}{Hong-Yu Zhou}.} \bibinfo{year}{2024}\natexlab{}.
\newblock \showarticletitle{Uncertainty Estimation of Large Language Models in Medical Question Answering}.
\newblock \bibinfo{journal}{\emph{arXiv preprint arXiv:2407.08662}} (\bibinfo{year}{2024}).
\newblock


\bibitem[Xing et~al\mbox{.}(2025)]%
        {xing2025re}
\bibfield{author}{\bibinfo{person}{Shuo Xing}, \bibinfo{person}{Yuping Wang}, \bibinfo{person}{Peiran Li}, \bibinfo{person}{Ruizheng Bai}, \bibinfo{person}{Yueqi Wang}, \bibinfo{person}{Chan-wei Hu}, \bibinfo{person}{Chengxuan Qian}, \bibinfo{person}{Huaxiu Yao}, {and} \bibinfo{person}{Zhengzhong Tu}.} \bibinfo{year}{2025}\natexlab{}.
\newblock \showarticletitle{Re-Align: Aligning Vision Language Models via Retrieval-Augmented Direct Preference Optimization}.
\newblock \bibinfo{journal}{\emph{arXiv preprint arXiv:2502.13146}} (\bibinfo{year}{2025}).
\newblock


\bibitem[Xiong et~al\mbox{.}(2024a)]%
        {xiong2024can}
\bibfield{author}{\bibinfo{person}{Miao Xiong}, \bibinfo{person}{Zhiyuan Hu}, \bibinfo{person}{Xinyang Lu}, \bibinfo{person}{YIFEI LI}, \bibinfo{person}{Jie Fu}, \bibinfo{person}{Junxian He}, {and} \bibinfo{person}{Bryan Hooi}.} \bibinfo{year}{2024}\natexlab{a}.
\newblock \showarticletitle{Can {LLM}s Express Their Uncertainty? An Empirical Evaluation of Confidence Elicitation in {LLM}s}. In \bibinfo{booktitle}{\emph{The Twelfth International Conference on Learning Representations}}.
\newblock


\bibitem[Xiong et~al\mbox{.}(2024b)]%
        {xiong2024efficient}
\bibfield{author}{\bibinfo{person}{Miao Xiong}, \bibinfo{person}{Andrea Santilli}, \bibinfo{person}{Michael Kirchhof}, \bibinfo{person}{Adam Golinski}, {and} \bibinfo{person}{Sinead Williamson}.} \bibinfo{year}{2024}\natexlab{b}.
\newblock \showarticletitle{Efficient and effective uncertainty quantification for LLMs}. In \bibinfo{booktitle}{\emph{Neurips Safe Generative AI Workshop 2024}}.
\newblock


\bibitem[Yang et~al\mbox{.}({[n.\,d.]})]%
        {yangbayesian}
\bibfield{author}{\bibinfo{person}{Adam~X Yang}, \bibinfo{person}{Maxime Robeyns}, \bibinfo{person}{Xi Wang}, {and} \bibinfo{person}{Laurence Aitchison}.} \bibinfo{year}{[n.\,d.]}\natexlab{}.
\newblock \showarticletitle{Bayesian Low-rank Adaptation for Large Language Models}. In \bibinfo{booktitle}{\emph{The Twelfth International Conference on Learning Representations}}.
\newblock


\bibitem[Yang et~al\mbox{.}(2025)]%
        {yang-etal-2025-maqa}
\bibfield{author}{\bibinfo{person}{Yongjin Yang}, \bibinfo{person}{Haneul Yoo}, {and} \bibinfo{person}{Hwaran Lee}.} \bibinfo{year}{2025}\natexlab{}.
\newblock \showarticletitle{{MAQA}: Evaluating Uncertainty Quantification in {LLM}s Regarding Data Uncertainty}. In \bibinfo{booktitle}{\emph{Findings of the Association for Computational Linguistics: NAACL 2025}}. \bibinfo{pages}{5846--5863}.
\newblock


\bibitem[Yang et~al\mbox{.}(2018)]%
        {yang2018hotpotqa}
\bibfield{author}{\bibinfo{person}{Zhilin Yang}, \bibinfo{person}{Peng Qi}, \bibinfo{person}{Saizheng Zhang}, \bibinfo{person}{Yoshua Bengio}, \bibinfo{person}{William~W Cohen}, \bibinfo{person}{Ruslan Salakhutdinov}, {and} \bibinfo{person}{Christopher~D Manning}.} \bibinfo{year}{2018}\natexlab{}.
\newblock \showarticletitle{HotpotQA: A dataset for diverse, explainable multi-hop question answering}.
\newblock \bibinfo{journal}{\emph{arXiv preprint arXiv:1809.09600}} (\bibinfo{year}{2018}).
\newblock


\bibitem[Yao et~al\mbox{.}(2025)]%
        {yao2025comal}
\bibfield{author}{\bibinfo{person}{Huaiyuan Yao}, \bibinfo{person}{Longchao Da}, \bibinfo{person}{Vishnu Nandam}, \bibinfo{person}{Justin Turnau}, \bibinfo{person}{Zhiwei Liu}, \bibinfo{person}{Linsey Pang}, {and} \bibinfo{person}{Hua Wei}.} \bibinfo{year}{2025}\natexlab{}.
\newblock \showarticletitle{Comal: Collaborative multi-agent large language models for mixed-autonomy traffic}. In \bibinfo{booktitle}{\emph{Proceedings of the 2025 SIAM International Conference on Data Mining (SDM)}}. \bibinfo{pages}{409--418}.
\newblock


\bibitem[Yao et~al\mbox{.}(2023)]%
        {yao2023tree}
\bibfield{author}{\bibinfo{person}{Shunyu Yao}, \bibinfo{person}{Dian Yu}, \bibinfo{person}{Jeffrey Zhao}, \bibinfo{person}{Izhak Shafran}, \bibinfo{person}{Tom Griffiths}, \bibinfo{person}{Yuan Cao}, {and} \bibinfo{person}{Karthik Narasimhan}.} \bibinfo{year}{2023}\natexlab{}.
\newblock \showarticletitle{Tree of thoughts: Deliberate problem solving with large language models}.
\newblock \bibinfo{journal}{\emph{Advances in neural information processing systems}}  \bibinfo{volume}{36} (\bibinfo{year}{2023}), \bibinfo{pages}{11809--11822}.
\newblock


\bibitem[Ye et~al\mbox{.}(2025)]%
        {ye2025benchmarking}
\bibfield{author}{\bibinfo{person}{Fanghua Ye}, \bibinfo{person}{Mingming Yang}, \bibinfo{person}{Jianhui Pang}, \bibinfo{person}{Longyue Wang}, \bibinfo{person}{Derek Wong}, \bibinfo{person}{Emine Yilmaz}, \bibinfo{person}{Shuming Shi}, {and} \bibinfo{person}{Zhaopeng Tu}.} \bibinfo{year}{2025}\natexlab{}.
\newblock \showarticletitle{Benchmarking llms via uncertainty quantification}.
\newblock \bibinfo{journal}{\emph{Advances in Neural Information Processing Systems}} (\bibinfo{year}{2025}), \bibinfo{pages}{15356--15385}.
\newblock


\bibitem[Ye et~al\mbox{.}(2024)]%
        {ye2024uncertainty}
\bibfield{author}{\bibinfo{person}{Kai Ye}, \bibinfo{person}{Tiejin Chen}, \bibinfo{person}{Hua Wei}, {and} \bibinfo{person}{Liang Zhan}.} \bibinfo{year}{2024}\natexlab{}.
\newblock \showarticletitle{Uncertainty regularized evidential regression}. In \bibinfo{booktitle}{\emph{Proceedings of the AAAI Conference on Artificial Intelligence}}, Vol.~\bibinfo{volume}{38}. \bibinfo{pages}{16460--16468}.
\newblock


\bibitem[Yin et~al\mbox{.}(2024)]%
        {yin2024reasoning}
\bibfield{author}{\bibinfo{person}{Zhangyue Yin}, \bibinfo{person}{Qiushi Sun}, \bibinfo{person}{Qipeng Guo}, \bibinfo{person}{Zhiyuan Zeng}, \bibinfo{person}{Xiaonan Li}, \bibinfo{person}{Junqi Dai}, \bibinfo{person}{Qinyuan Cheng}, \bibinfo{person}{Xuan-Jing Huang}, {and} \bibinfo{person}{Xipeng Qiu}.} \bibinfo{year}{2024}\natexlab{}.
\newblock \showarticletitle{Reasoning in flux: Enhancing large language models reasoning through uncertainty-aware adaptive guidance}. In \bibinfo{booktitle}{\emph{Proceedings of the 62nd Annual Meeting of the Association for Computational Linguistics (Volume 1: Long Papers)}}. \bibinfo{pages}{2401--2416}.
\newblock


\bibitem[Young et~al\mbox{.}(2025)]%
        {young2024flexible}
\bibfield{author}{\bibinfo{person}{Spencer Young}, \bibinfo{person}{Porter Jenkins}, \bibinfo{person}{Lonchao Da}, \bibinfo{person}{Jeff Dotson}, {and} \bibinfo{person}{Hua Wei}.} \bibinfo{year}{2025}\natexlab{}.
\newblock \showarticletitle{Flexible heteroscedastic count regression with deep double poisson networks}.
\newblock \bibinfo{journal}{\emph{International Conference on Machine Learning}} (\bibinfo{year}{2025}).
\newblock


\bibitem[Zamani et~al\mbox{.}(2020)]%
        {zamani2020generating}
\bibfield{author}{\bibinfo{person}{Hamed Zamani}, \bibinfo{person}{Susan Dumais}, \bibinfo{person}{Nick Craswell}, \bibinfo{person}{Paul Bennett}, {and} \bibinfo{person}{Gord Lueck}.} \bibinfo{year}{2020}\natexlab{}.
\newblock \showarticletitle{Generating clarifying questions for information retrieval}. In \bibinfo{booktitle}{\emph{Proceedings of the web conference 2020}}. \bibinfo{pages}{418--428}.
\newblock


\bibitem[Zellers et~al\mbox{.}(2019)]%
        {zellers2019hellaswag}
\bibfield{author}{\bibinfo{person}{Rowan Zellers}, \bibinfo{person}{Ari Holtzman}, \bibinfo{person}{Yonatan Bisk}, \bibinfo{person}{Ali Farhadi}, {and} \bibinfo{person}{Yejin Choi}.} \bibinfo{year}{2019}\natexlab{}.
\newblock \showarticletitle{HellaSwag: Can a Machine Really Finish Your Sentence?}. In \bibinfo{booktitle}{\emph{Proceedings of the 57th Annual Meeting of the Association for Computational Linguistics}}. \bibinfo{pages}{4791--4800}.
\newblock


\bibitem[Zhang and Zhang(2025)]%
        {zhang2025cot}
\bibfield{author}{\bibinfo{person}{Boxuan Zhang} {and} \bibinfo{person}{Ruqi Zhang}.} \bibinfo{year}{2025}\natexlab{}.
\newblock \showarticletitle{CoT-UQ: Improving Response-wise Uncertainty Quantification in LLMs with Chain-of-Thought}.
\newblock \bibinfo{journal}{\emph{arXiv preprint arXiv:2502.17214}} (\bibinfo{year}{2025}).
\newblock


\bibitem[Zhang et~al\mbox{.}(2024b)]%
        {zhang2024luq}
\bibfield{author}{\bibinfo{person}{Caiqi Zhang}, \bibinfo{person}{Fangyu Liu}, \bibinfo{person}{Marco Basaldella}, {and} \bibinfo{person}{Nigel Collier}.} \bibinfo{year}{2024}\natexlab{b}.
\newblock \showarticletitle{LUQ: Long-text Uncertainty Quantification for LLMs}. In \bibinfo{booktitle}{\emph{Proceedings of the 2024 Conference on Empirical Methods in Natural Language Processing}}. \bibinfo{pages}{5244--5262}.
\newblock


\bibitem[Zhang et~al\mbox{.}(2024d)]%
        {zhang2024vl}
\bibfield{author}{\bibinfo{person}{Ruiyang Zhang}, \bibinfo{person}{Hu Zhang}, {and} \bibinfo{person}{Zhedong Zheng}.} \bibinfo{year}{2024}\natexlab{d}.
\newblock \showarticletitle{VL-Uncertainty: Detecting Hallucination in Large Vision-Language Model via Uncertainty Estimation}.
\newblock \bibinfo{journal}{\emph{arXiv preprint arXiv:2411.11919}} (\bibinfo{year}{2024}).
\newblock


\bibitem[Zhang et~al\mbox{.}({[n.\,d.]})]%
        {zhangbertscore}
\bibfield{author}{\bibinfo{person}{Tianyi Zhang}, \bibinfo{person}{Varsha Kishore}, \bibinfo{person}{Felix Wu}, \bibinfo{person}{Kilian~Q Weinberger}, {and} \bibinfo{person}{Yoav Artzi}.} \bibinfo{year}{[n.\,d.]}\natexlab{}.
\newblock \showarticletitle{BERTScore: Evaluating Text Generation with BERT}. In \bibinfo{booktitle}{\emph{International Conference on Learning Representations}}.
\newblock


\bibitem[Zhang et~al\mbox{.}(2024a)]%
        {zhang2024unveiling}
\bibfield{author}{\bibinfo{person}{Yuan Zhang}, \bibinfo{person}{Tao Huang}, \bibinfo{person}{Chun-Kai Fan}, \bibinfo{person}{Hongyuan Dong}, \bibinfo{person}{Jiawen Li}, \bibinfo{person}{Jiacong Wang}, \bibinfo{person}{Kuan Cheng}, \bibinfo{person}{Shanghang Zhang}, \bibinfo{person}{Haoyuan Guo}, {et~al\mbox{.}}} \bibinfo{year}{2024}\natexlab{a}.
\newblock \showarticletitle{Unveiling the tapestry of consistency in large vision-language models}.
\newblock \bibinfo{journal}{\emph{Advances in Neural Information Processing Systems}}  \bibinfo{volume}{37} (\bibinfo{year}{2024}), \bibinfo{pages}{118632--118653}.
\newblock


\bibitem[Zhang et~al\mbox{.}(2024c)]%
        {zhang2024badcm}
\bibfield{author}{\bibinfo{person}{Zheng Zhang}, \bibinfo{person}{Xu Yuan}, \bibinfo{person}{Lei Zhu}, \bibinfo{person}{Jingkuan Song}, {and} \bibinfo{person}{Liqiang Nie}.} \bibinfo{year}{2024}\natexlab{c}.
\newblock \showarticletitle{BadCM: Invisible backdoor attack against cross-modal learning}.
\newblock \bibinfo{journal}{\emph{IEEE Transactions on Image Processing}} (\bibinfo{year}{2024}).
\newblock


\bibitem[Zhao et~al\mbox{.}(2023b)]%
        {zhao2023radiology}
\bibfield{author}{\bibinfo{person}{Guosheng Zhao}, \bibinfo{person}{Zijian Zhao}, \bibinfo{person}{Wuxian Gong}, {and} \bibinfo{person}{Feng Li}.} \bibinfo{year}{2023}\natexlab{b}.
\newblock \showarticletitle{Radiology report generation with medical knowledge and multilevel image-report alignment: A new method and its verification}.
\newblock \bibinfo{journal}{\emph{Artificial Intelligence in Medicine}}  \bibinfo{volume}{146} (\bibinfo{year}{2023}), \bibinfo{pages}{102714}.
\newblock


\bibitem[Zhao et~al\mbox{.}(2023a)]%
        {zhao2023calibrating}
\bibfield{author}{\bibinfo{person}{Yao Zhao}, \bibinfo{person}{Mikhail Khalman}, \bibinfo{person}{Rishabh Joshi}, \bibinfo{person}{Shashi Narayan}, \bibinfo{person}{Mohammad Saleh}, {and} \bibinfo{person}{Peter~J Liu}.} \bibinfo{year}{2023}\natexlab{a}.
\newblock \showarticletitle{Calibrating Sequence likelihood Improves Conditional Language Generation}. In \bibinfo{booktitle}{\emph{The Eleventh International Conference on Learning Representations}}.
\newblock


\bibitem[Zheng et~al\mbox{.}(2023)]%
        {zheng2023large}
\bibfield{author}{\bibinfo{person}{Chujie Zheng}, \bibinfo{person}{Hao Zhou}, \bibinfo{person}{Fandong Meng}, \bibinfo{person}{Jie Zhou}, {and} \bibinfo{person}{Minlie Huang}.} \bibinfo{year}{2023}\natexlab{}.
\newblock \showarticletitle{Large language models are not robust multiple choice selectors}.
\newblock \bibinfo{journal}{\emph{arXiv preprint arXiv:2309.03882}} (\bibinfo{year}{2023}).
\newblock


\bibitem[Zheng et~al\mbox{.}(2024)]%
        {zheng2024evaluating}
\bibfield{author}{\bibinfo{person}{Zhi Zheng}, \bibinfo{person}{Qian Feng}, \bibinfo{person}{Hang Li}, \bibinfo{person}{Alois Knoll}, {and} \bibinfo{person}{Jianxiang Feng}.} \bibinfo{year}{2024}\natexlab{}.
\newblock \showarticletitle{Evaluating uncertainty-based failure detection for closed-loop llm planners}.
\newblock \bibinfo{journal}{\emph{arXiv preprint arXiv:2406.00430}} (\bibinfo{year}{2024}).
\newblock


\end{thebibliography}










\end{document}